\pdfoutput=1

\documentclass[11pt]{article}

\usepackage[preprint]{acl}

\usepackage{times}
\usepackage{latexsym}
\usepackage{subcaption}

\usepackage[T1]{fontenc}

\usepackage[utf8]{inputenc}

\usepackage{microtype}

\usepackage{inconsolata}

\usepackage{graphicx}

\usepackage{url}
\usepackage{multirow}
\usepackage{booktabs}
\usepackage{amsmath}
\usepackage{todonotes}
\usepackage{amssymb}
\usepackage{enumitem}
\usepackage{makecell}
\usepackage{paralist}

\title{Show Me How You Reason and I'll Tell You Who You Are: \\ Reasoning Graphs for Robust LLM Authorship Attribution}

\author{
  \textbf{Zlata Kikteva\textsuperscript{1}},
  \textbf{Artur Romazanov\textsuperscript{1}},
  \textbf{Annette Hautli-Janisz\textsuperscript{1}}, and
  \textbf{Ramon Ruiz-Dolz\textsuperscript{2}}
\\
  \textsuperscript{1}University of Passau, Germany \\
  \textsuperscript{2}University of Dundee, United Kingdom 
}

\begin{document}
\maketitle
\begin{abstract}
Given the current trend to employ large language models (LLMs) in almost any imaginable context, LLM-generated text detection and authorship attribution have become a pressing issue. Prior work has primarily focused on surface-level linguistic features, an approach shown to be susceptible to paraphrasing and other obfuscation techniques. In this paper, we go beyond the linguistic surface, extracting and analysing reasoning structures in LLM-generated texts with the goal of capturing more complex signals of LLM authorship. We propose a graph neural network approach that leverages reasoning graphs extracted by an argument mining pipeline, demonstrating improved robustness and generalisation over a traditional Longformer baseline. Our approach outperforms the baseline by up to 27 percentage points under the obfuscation attacks such as paraphrasing and backtranslation, and 19 percentage points when evaluated on the texts generated by the unseen model versions, simulating real-world conditions in which new LLM versions are continuously released. 

\end{abstract}

\section{Introduction}

Ever since the popularisation of large language models (LLMs) and their growing ubiquity in 
education, scientific writing, and peer review, they have the potential for being misused. 
As \citet{kasneci2023chatgpt} observe in the education context, ``\textit{It is becoming increasingly difficult to distinguish whether a text is machine- or human-generated, presenting an additional major challenge to teachers and educators} \cite{cotton2023chatting, elkins2020can, gao2022comparing, dehouche2021plagiarism}.'' The matter is further illustrated in science by ICLR-26 suspecting 21\% of fully AI-generated paper reviews.\footnote{\url{https://www.nature.com/articles/d41586-025-03506-6}} Simultaneously, there are growing concerns among the general public over the proportion of LLM-generated to human-authored content online,\footnote{\url{https://graphite.io/five-percent/ai-now-writes-as-many-online-articles-as-humans-do}} raising questions about the future of the internet.\footnote{\url{https://www.forbes.com.au/news/innovation/is-ai-quietly-killing-itself-and-the-internet}} Indeed, researchers report on increased proliferation of LLM-authored material on social media and other online spheres \cite{sun-etal-2025-ai, dolezal2026impact}. As such, there is no denying the fact of a significant rise in machine authorship in the near past. 

LLM-authored text detection task has generated considerable interest within the NLP community, making it one of the prominent research questions in recent years \cite{wu2025survey}. This is further evidenced by the significant number of machine-generated text detection benchmarks and datasets covering multiple languages, domains, and generative models \cite{macko2023multitude, la2025openturingbench}. With numerous studies reporting on the high performance of detection models \cite{wang2024m4gt, mitchell2023detectgpt, hans2024spotting}, the task might appear trivial. However, most of the benchmarks focus on specific model versions and rarely consider generalisation capabilities across versions, a scenario which reflects the challenges of a fast-paced technological landscape. 

Furthermore, various authorship obfuscation techniques, such as word-level perturbations or text paraphrasing, have been shown to noticeably decrease the detection models' performance \cite{macko2024authorship, zhou2024navigating, xing2024alison}. With the proliferation of tools for `humanising' AI-authored text online, it is necessary to include obfuscated text variants in the evaluation of detection models if the aim is to assess performance in a setting that reflects real-world use of AI detection systems, e.g., educators verifying authorship in students' writing. 

\begin{figure*}[t]
    \centering
    \includegraphics[width=1\textwidth]{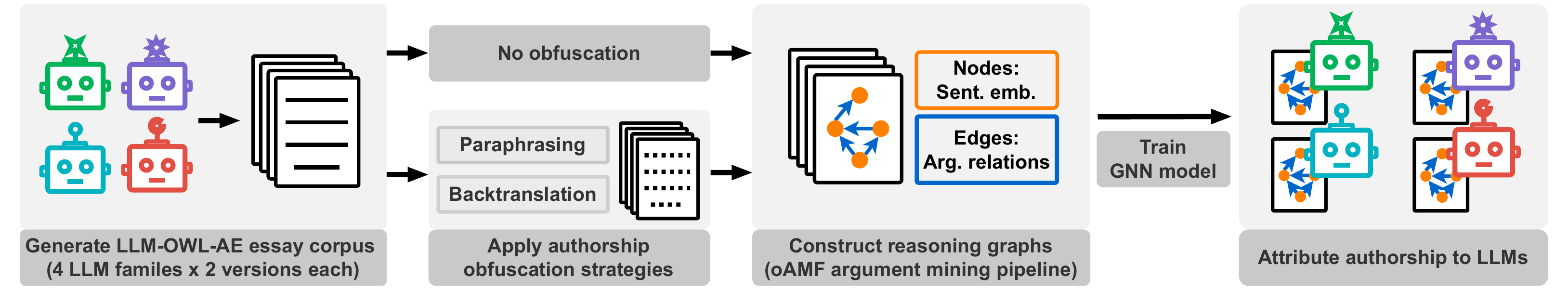}
    \caption{Reasoning graph-based approach using GNN architecture for LLM authorship attribution. 
    }
    \label{fig:method}
\end{figure*}

In other scenarios, however, it is necessary to go deeper than simple detection. Earlier work suggests that models from different providers exhibit different behavioural tendencies in terms of their biases \cite{bang2024measuring}, ideologies \cite{buyl2026large}, or capabilities for disinformation generation \cite{vykopal-etal-2024-disinformation}. Combined with the risk of misuse of the LLM technology, for example, for the automation of online disinformation campaigns \cite{guo2024online}, it is increasingly important to be able to identify the model that authored a text. Some work has begun to move beyond the problem of detecting LLM-generated text \cite{la2025openturingbench}, focusing on \textit{authorship attribution}, i.e., backtracking text's origin to a specific model provider \cite{yang2023dna}. In this paper, we address the task of attributing authorship of LLM-generated texts, with the challenges that arise in real-world application scenarios in mind. To this end, we focus on two key challenges: (1) \textit{Robustness} in conditions when the LLM-authored texts are obfuscated, and (2) \textit{generalisation} across different versions of LLMs belonging to the same family.

Prior work has shown that with only surface-level lexical features as an input, it is often difficult (or impossible) to ascertain the specific model version that was used to generate a text \cite{huang2025authorship}. In addition to that, \citet{li-etal-2025-prdetect} demonstrate that a syntax-graph-based approach exhibits a much more robust performance under perturbation conditions than current transformer-based approaches. Given these insights, we propose a novel approach to authorship attribution with reasoning graphs as input to train a classification model, using generalisation and robustness as the two main evaluation criteria. 



By learning from structured graph-based representations of the underlying reasoning, constructed using an argument mining pipeline, our model predicts the model that was used to generate a text based on structural reasoning features in the form of arguments and the relations between them. 
By way of graph neural networks (GNNs), models that have been applied to the argumentation-related tasks, including argument acceptability prediction \cite{kuhlmann2019using,malmqvist2020determining, cibier2024graph, gehlot2026heterogeneous}, argumentative component identification \cite{ruggeri2021tree, Zhang-arg}, matching argument extraction \cite{mao2024seeing}, and winning stance prediction \cite{ruiz2023automatic}, we go significantly beyond linguistic and semantic representations for authorship attribution and build on a meta-linguistic, pragmatic representation of the underlying textual structure. To evaluate the robustness and generalisation of the reasoning graph-based approach, we create a corpus of 2,240 LLM-generated argumentative essays spanning eight different model versions belonging to four different families (Gemma, Qwen, Llama, and Phi). We further augment this with three obfuscated variants per essay via paraphrasing and backtranslation (through French and Turkish), resulting in a corpus of 8,960 essays in total. Our approach to authorship attribution with reasoning graphs is visualised in Figure \ref{fig:method}.



Results show that 
our reasoning graph-based fingerprinting fares significantly better (up to +27\% F1-score) than the text-only baseline in terms of both robustness (i.e., under the obfuscation conditions) and generalisation, when we train and test on different model versions of the same family (e.g., train on Llama 3.3 and test on Llama 4). Our contributions are as follows: (1) We create and publicly release a new dataset of LLM-generated argumentative essays across different model versions and families for generalisation assessment and under different perturbation strategies for robustness assessment. (2) We propose a novel reasoning graph-based approach to authorship attribution, combining language and reasoning-based features, which demonstrates superior performance in terms of robustness and generalisation when compared to a text-only approach.




    



\section{Related Work}

\paragraph{Detection} \citet{liu2023argugpt} investigate the task of LLM-generated text detection for argumentative essays authored by different GPT models and report high performance results on the task. Such findings are 
not uncommon in Transformer-based approaches to LLM-generated text detection under in-domain evaluation scenarios \cite{wang2024m4gt}. In contrast, \citet{abassy2024llm} report a noticeable drop in performance when evaluating on the unseen domains and generators, with \citet{dugan2024raid} making similar observations. Furthermore, the detector model accuracies vary when evaluated on the different model versions from the same family \cite{yu2025evobench}. This highlights a critical challenge: 90\%+ accuracy has limited practical significance if the performance does not generalise well, 
especially given the fast-paced landscape where new models and model versions are released regularly. 

\paragraph{Obfuscation}


At the same time, some research focuses on the robustness of the detection models when faced with the intentionally modified versions of the LLM-generated texts, aiming to investigate the extent to which the obfuscation hinders the detection and attribution of generative models \cite{uchendu2023attribution, chakraborty-etal-2023-counter, macko2024authorship}. Some of the most widely used approaches rely on rewriting or paraphrasing, either by way of LLMs \cite{fang2025your} or dedicated paraphrasing tools \cite{krishna2023paraphrasing, sadasivan2023can}, as well as backtranslation \cite{altakrori-etal-2022-multifaceted, macko2024authorship, ayoobi2025esperanto}, while word-level perturbations are frequently employed in the adversarial attacks targeting the detector architecture \cite{shi2024red, wang2024raft, zhou2024humanizing}. 

\paragraph{Attribution} The significant increase in model providers, each of them exhibiting different generative styles, has sparked interest in a particular sub-task: LLM authorship attribution \cite{yang2023dna}. Beyond simply identifying whether a text is human-written or LLM-generated (or a mixture of both), authorship attribution aims at identifying the specific model that generated a given text \cite{li2023origin}. One of the first studies in this area conducts a stylometric analysis based on lexical, syntactic, and structural features of texts generated by six different models from the GPT and Llama families \cite{kumarage2023neural}. Their findings reveal clear differences between closed and open-weight model families, evidencing the relevance of analysing models not only individually but also considering their families. 
A recent analysis highlights the challenges of LLM authorship attribution when relying only on surface-level language features, especially given how fast generative models change and improve \cite{bevendorff2025two}. The approach adopted in this paper goes beyond the surface-based line of earlier work by taking into account the underlying reasoning structure of essays in order to attribute individual texts to specific LLMs and their versions. 



\section{Data}

We create a new corpus of LLM-generated argumentative essays based on the Purdue Online Writing Lab guidelines,\footnote{\url{https://owl.purdue.edu/owl/general\_writing/academic\_writing/essay\_writing/argumentative\_essays.html}} a widely used academic writing resource. There are two main reasons for creating our own dataset: First, it is necessary to ensure the argumentative essay length is sufficient for constructing informative reasoning graphs. 
This condition rules out a significant amount of publicly available corpora, because these consist of short machine-generated text sequences. Secondly, to evaluate the generalisation of our approach,
data generated by LLMs belonging to different families with at least two model versions per family is required -- a condition that none of the earlier datasets fulfills. In the following sections, we describe the process to generate 
LLM-OWL-AE.\footnote{The LLM-OWL-AE corpus will be released under a Creative Commons Attribution-NonCommercial-ShareAlike 4.0 International licence (CC BY-NC-SA 4.0) and will be made publicly available in a GitHub repository together with the code upon acceptance of the paper.} 


\subsection{Data Generation}

The generation process covers eight LLMs from four different families (two versions per family): Gemma3 (27B), Gemma4 (31B), Qwen3 (32B), Qwen3.5 (35B), Llama3.3 (70B), Llama4 (109B), Phi3 (14B), and Phi4 (14B). 
We refer to the essays generated by the earlier model versions (Gemma3 released in Mar. 2025, Qwen3 in Apr. 2025, Llama3.3 and Phi3 in Apr. 2024) as LLM-OWL-AE-I. The essays generated with the later model versions are called LLM-OWL-AE-II (Gemma4 released in Mar. 2026, Qwen3.5 in Feb. 2026, Llama4 in Apr. 2025, Phi4 in Dec. 2024). We focus on open LLMs because, as pointed out in previous work, these models make the task more challenging due to the increased diversity in their architectures and training data \cite{la2025openturingbench}. All models are run with their respective 4-bit quantised versions in Ollama\footnote{\url{https://ollama.com/}} with temperature set to 1. They were prompted to generate essays of about 500 words (max. 600, min. 400) following the Purdue OWL argumentative essay writing instructions, covering 140 different topics, such as animal rights or climate change, and two stances (one in favour and one against) for each topic. Topics are adopted from \citet{ruiz2024nlas}. The process results in a collection of 2,240 essays (280 essays per model version; 560 essays per model family).

\subsection{Obfuscated Data Variants}
To investigate the robustness of the models' performance on the task of authorship attribution, we adopt two widely used authorship obfuscation techniques, namely paraphrasing and backtranslation \cite{macko2024authorship, zhou2024navigating}. These methods introduce lexical and syntactic variation while preserving semantic meaning, thus potentially reducing authorship signals. We do not employ word-level substitution strategies 
or character-level perturbations. The former would only have a limited impact on the reasoning graph, while the latter might negatively impact the performance of the argument identification pipeline. 

\paragraph{Paraphrasing} For paraphrasing, we utilise DIPPER paraphrasing tool \cite{krishna2023paraphrasing}, a detection-model-agnostic method (as opposed to the adversarial paraphrasers) that performs lexical rephrasing as well as sentence reordering. As DIPPER is trained to rewrite text at the paragraph level, paraphrasing is applied to each paragraph individually, and then the paraphrased paragraphs are recombined in their original order to reconstruct the full essay. For lexical and order diversity parameters, the settings are adopted as reported by the authors of the paper to achieve optimal performance (L60, O60). 

\paragraph{Backtranslation} Open-source machine-translation models by OPUS-MT \cite{tiedemann-thottingal-2020-opus, tiedemann-2020-tatoeba} are used for backtranslation via French and Turkish languages. We choose French as a lexically and syntactically close language to English. Turkish is chosen due to its minimal lexical overlap with English and its distinct morphological properties as an agglutinative language. This allows us to test different degrees of obfuscation. For consistency with the paraphrasing setup, backtranslation is also performed at the paragraph level. We conduct a manual sanity check on a sample of the data to ensure that the obfuscated texts remain semantically coherent and preserve the meaning of the original texts.


\subsection{Essay Linguistic Surface}

In terms of the linguistic surface of the generated essays, we find several differences across different model families, model versions, as well as different perturbation strategies, indicating a varied linguistic surface learned during the training of a text-based model. The detailed dataset statistics are presented in Appendix \ref{app:ling_stats}.

\paragraph{Essay length} Depending on the model, the average length of an essay ranges from 466 to 543 words, with later model versions generating shorter essays. When it comes to the perturbation, backtranslation produces shorter essays compared to the original ones, with an average reduction of 140 words for French and 80 words for Turkish, while paraphrasing tends to maintain the original essay length better. 

\paragraph{Lexical diversity} The lexical diversity of the essays is measured with the MTLD score \cite{mccarthy2010mtld} as an approximation of the diversity of the used vocabulary. We find that both model versions of the Llama family generate essays with the least diverse vocabulary, indicated by the scores of 103 and 87, while other models have scores ranging from 129 to 196 (higher scores indicate a higher degree of lexical diversity). Unlike the word count, the MTLD scores appear to be maintained better between the model versions. Paraphraser has a negative effect on the richness of the vocabulary (scores ranging from 72 to 96), which results in models of the Qwen and Phi families dropping about 90 or more points when compared to the original essays. The backtranslation strategy has less of a negative impact on the lexical diversity, with the scores dropping by no more than 50 points.

\paragraph{Syntactic complexity} The syntactic complexity of the essays is approximated in terms of the mean number of conjuncts, clausal modifiers of nouns, adverbial clause modifiers, clausal complements, clausal subjects and parataxis per sentence. Models of the Qwen family, especially the later version, tend to generate simpler sentences, while earlier Llama and Gemma use more complex structures. Under the paraphrasing strategy and backtranslation via Turkish, there is a minor drop in complexity, with sentences after French backtranslation being more complex than the original ones. 







\section{Method}
\label{sec:method}



Our approach to authorship attribution is as follows: We first apply an existing argument mining pipeline to extract argument relations within each essay, representing each essay as a graph with text segments as nodes and the argument relations as edges. The resulting graphs are used to train and evaluate a GNN, which classifies each essay into model families and versions. 


\paragraph{Argument relation extraction} oAMF (Open Argument Mining Framework) \cite{gemechu-etal-2025-open} is an open-source modular end-to-end argument mining pipeline. The TARGER module \cite{chernodub-etal-2019-targer} segments the individual essays into argumentative discourse units (ADUs), which roughly correspond to sentences. We then use the ARIR module \cite{ruiz2021transformer} to predict the probability of inference (relation of support), conflict (relation of attack), rephrase (relation between two ADUs when one is used to reformulate another), and no relation between two ADUs. 

\paragraph{Graph construction} As the first step in the pipeline, we create graph representation $G=(V,E,W)$, in which nodes $V=\{v_i\}$ are ordered according to their appearance in the text, and directed edges $E = \{(v_i, v_j)\ \in V \times V \mid i<j\}$ represent relations $W:E\to[0,1]^4, W(e)=(w_\mathrm{inf}, w_\mathrm{conf}, w_\mathrm{rephr},w_\mathrm{no\_rel}),\sum_{k} w_k(e) = 1$ between nodes as retrieved from the ARIR module. 
The existence of an edge is verified 
with predefined thresholds $T=(t_{inf}, t_{conf}, t_{rephr})$ for relations of inference, conflict, and rephrase. If any probability of the relation is above the threshold for the respective relation, we retain the edge $E'=\{e \in E \;\;\exists k \in \{\mathrm{inf},\mathrm{conf},\mathrm{rephr}\}:w_k(e)\ge T_k\}$. 

\paragraph{Node encoding} The sentence-transformer model all-mpnet-base-v2 \cite{reimers-2019-sentence-bert} generates the sentence embeddings for each graph node $v_i$. 

\paragraph{Edge encoding}
There are two interchangeable edge processing strategies. Under the \textit{argmax} strategy, the raw probability distribution over the relation classes from the ARIR module is converted into a one-hot-encoded vector by assigning a value of 1 to the class with the highest probability and 0 to all others $W_{\mathrm{argmax}}=(w_\mathrm{inf}, w_\mathrm{conf}, w_\mathrm{rephr})$. The resulting vector contains three elements, each corresponding to one of the relation classes: inference, conflict, and rephrase. With the \textit{probabilities} strategy, the four-element vector contains raw probabilities from the ARIR module for the three relation classes (inference, conflict, rephrase) and the no relation probability $W_{\mathrm{probs}}=(w_\mathrm{inf}, w_\mathrm{conf}, w_\mathrm{rephr},w_\mathrm{no\_rel})$. Under each of the edge processing strategies, a separate homogeneous directed graph is constructed for each essay.


\paragraph{Graph Classification}The resulting graphs are being classified by a $\mathrm{GNN}:\mathcal{G} \to \mathcal{Y}$, which aggregates the node representations of a graph to predict a label for a model $\quad \hat{y}=\mathrm{GNN}(G)$ that produced an original essay.




\section{Experiments}
\label{sec:experiments}
\begin{table}
\centering
\small
\caption{Train/test splits across dataset partitions. I refers to data partition LLM-OWL-AE-I; II to LLM-OWL-AE-II. The \textsc{Same-Version} setting includes an extended test set with 80 additional essays per model, indicated by a * next to the data partition name.}
\label{tab:eval-config}

\begin{tabular}{lcc}
\toprule
Config & Train set & Test set \\
\midrule

\multirow{2}{*}{\textsc{Same-Version}}
& I  & I*  \\
& II & II* \\

\midrule

\multirow{2}{*}{\textsc{Cross-Version}}
& I  & II \\
& II & I  \\

\bottomrule
\end{tabular}

\end{table}

\subsection{Experimental Design} 
\paragraph{Train/test splits} We consider two train/test split configurations of the LLM-OWL-AE corpus: \textsc{Same-Version} is a standard setup, in which both training and test data are generated by the same model version; \textsc{Cross-Version} evaluates generalisation to the data produced by unseen model version. 

As illustrated in Table \ref{tab:eval-config}, in \textsc{Same-Version}, the two dataset partitions, LLM-OWL-AE-I (essays by earlier model versions, i.e., Gemma3, Qwen3, Llama3.3, Phi3) and LLM-OWL-AE-II (essays by later model versions, i.e., Gemma4, Qwen3.5, Llama4, Phi4), are treated independently, i.e., for each partition, training, development and test data are drawn from the same partition, e.g., models trained on LLM-OWL-AE-I are evaluated on LLM-OWL-AE-I. We use 200 essays per model for training and development, and an additional 80 essays per model for testing, covering 100 topics in the training set and 40 topics in the test.  

In \textsc{Cross-Version}, the models are trained and evaluated across the partition: The models trained on the LLM-OWL-AE-I are evaluated on the LLM-OWL-AE-II, and the models trained on LLM-OWL-AE-II are evaluated on LLM-OWL-AE-I. We use 200 essays per model for training from one partition and 200 essays per model for testing from the other partition. 

\paragraph{Obfuscation strategies} To investigate \textit{robustness}, we evaluate the models' performance on the original essays (i.e., non-obfuscated essays) and essays obfuscated under paraphrasing, backtranslation via French, and backtranslation via Turkish strategies. 

\paragraph{Edge processing strategies in graphs}
We apply thresholds to the relation probabilities from the ARIR module to filter the edges with low label certainty. To evaluate how different decisions boundary for the argument relation prediction affect GNN models' performance, we define the three sets of thresholds $T_1=(t_{\mathrm{inf}}^{(1)},t_{\mathrm{conf}}^{(1)},t_{\mathrm{rephr}}^{(1)})=(0.5,0.5,0.5)$, $T_2=(t_{\mathrm{inf}}^{(2)},t_{\mathrm{conf}}^{(2)},t_{\mathrm{rephr}}^{(2)})=(0.7,0.7,0.7)$, and $T_3=(t_{\mathrm{inf}}^{(3)},t_{\mathrm{conf}}^{(3)},t_{\mathrm{rephr}}^{(3)})=(0.9,0.7,0.7)$. This means that, for instance, when threshold $T_1$ is applied, all edges with relation probability for inference, rephrase, and conflict below 0.5 are dropped. Each threshold configuration results in a separate set of graphs. Additionally, we create a set of graphs with the thresholds set at 0, thus retaining all of the edges. Edge processing strategies, \textit{argmax} and \textit{probabilities}, are then applied to the generated graphs, transforming their edges, to produce the final graph structures that are ready for classification. This process results in eight graphs for each essay: two edge processing strategies (\textit{argmax} and \textit{probabilities}) x four threshold strategies (none, $T_1$, $T_2$, $T_3$).

\subsection{Baseline} We consider a text-only baseline that models authorship attribution as a text sequence classification problem. For that purpose, we make use of the Longformer architecture \cite{beltagy2020longformer}, which extracts semantically rich features from long text input documents. The model is trained for 5 epochs on the four-class classification task, with a learning rate of 1e-5 and a 0.01 weight decay. Each experiment is repeated with 3 random seeds, and we report the average F1-score across the runs in Tables \ref{tab:samemodel-versions} and \ref{tab:diffmodel-versions} (\textit{Text-only baseline}).

\subsection{Graph Neural Networks}

Reasoning graphs are used to train and evaluate four distinct GNN layer types: Graph Convolutional Networks (GCN) \cite{Kipf-GCN}, Graph Attention Networks (GAT) \cite{velikovi2017graph}, Graph Transformer \cite{shi2021masked}, and General Powerful Scalable (GPS) networks \cite{rampavsek2022recipe}. The architectures differ primarily in their neighbourhood aggregation mechanisms, with GCN as a baseline using averaged neighbour aggregation, GAT learning attention weights for the neighbours, Graph Transformer capturing global attention, and GPS combining local message-passing with transformer-style global attention. Based on the average path length of 7.5 for the essay graphs, we evaluate configurations with up to 7 GNN layers as an empirically plausible range \cite{Li_Han_Wu_2018}. 

Node representations from the GNN layers of a model are aggregated into a graph-level embedding using global attention pooling. The resulting representation is passed to a classification head consisting of a LayerNorm, a 256-dimensional feed-forward layer with GELU activation, dropout with rate 0.2, and a final feed-forward projection to the label space of the 4 families: Gemma, Qwen, Llama, or Phi $\mathrm{GNN}:\mathcal{G} \to \mathcal{Y}, \quad \hat{y}=\mathrm{GNN}(G), \quad \mathcal{Y}=\{\mathrm{Gemma}, \mathrm{Qwen}, \mathrm{Llama}, \mathrm{Phi}\}$. All models are trained for 30 epochs with early stopping (patience = 5). Each experiment is repeated with 3 random seeds, and we report the average F1-score across the runs in Tables \ref{tab:samemodel-versions} and \ref{tab:diffmodel-versions} (\textit{Text + Reasoning Structure}).


\begin{table*}[t]
\small
\centering
\caption{Macro F1-score results in the \textsc{Same-Version} setup. \textit{None} in Threshold column indicates a complete graph; \textit{best} refers to the best-performing threshold among $T_1, T_2, T_3$, which are defined in Section \ref{sec:experiments}. \textit{Argmax} and \textit{probs} refer to the edge processing strategies defined in Section \ref{sec:method}. `Orig.' refers to the non-obfuscated essays, `paraphr.' to the paraphrased, `BT-FR' and `BT-TR' to the backtranslated essays via French and Turkish, respectively.}
\label{tab:samemodel-versions}

\begin{tabular}{lllcccccccc}
\toprule
& &
& \multicolumn{4}{c}{LLM-OWL-AE-I}
& \multicolumn{4}{c}{LLM-OWL-AE-II} \\
\cmidrule(lr){4-7} \cmidrule(lr){8-11}

Model & Edge & Thresh.
& Orig. & Paraphr. & BT-FR & BT-TR
& Orig. & Paraphr. & BT-FR & BT-TR \\
\midrule

\multicolumn{11}{l}{\textit{Text-only baseline}} \\
Longformer          & -- & -- & \textbf{0.97} & 0.63 & \textbf{0.84} & \textbf{0.88} &\textbf{ 0.96} & 0.45 & 0.62 & 0.69 \\
\midrule

\multicolumn{11}{l}{\textit{Text + Reasoning Structure}} \\
\multirow[t]{4}{*}{GCN} & \multirow[t]{2}{*}{argmax} & none 
 & 0.66 & 0.56 & 0.57 & 0.61 & 0.68 & 0.59 & 0.61 & 0.64 \\
 &  & best                                                
 & 0.69 & 0.59 & 0.58 & 0.66 & 0.72 & 0.63 & 0.62 & 0.65 \\
 & \multirow[t]{2}{*}{probs} & none                         
 & 0.66 & 0.56 & 0.57 & 0.61 & 0.68 & 0.59 & 0.61 & 0.64 \\
 &  & best                                                
 & 0.67 & 0.59 & 0.58 & 0.66 & 0.72 & 0.63 & 0.62 & 0.65 \\
\midrule
\multirow[t]{4}{*}{GAT} & \multirow[t]{2}{*}{argmax} & none 
 & 0.73 & 0.60 & 0.67 & 0.66 & 0.75 & 0.61 & 0.68 & 0.70 \\
 &  & best                                                
 & 0.71 & 0.59 & 0.65 & 0.68 & 0.75 & 0.64 & 0.67 & 0.70 \\
 & \multirow[t]{2}{*}{probs} & none                         
 & 0.74 & 0.62 & 0.66 & 0.67 & 0.77 & 0.63 & 0.67 & \textbf{0.72} \\
 &  & best                                                
 & 0.69 & 0.63 & 0.65 & 0.66 & 0.76 & 0.61 & 0.65 & 0.70 \\
\midrule
\multirow[t]{4}{*}{Graph Transformer} & \multirow[t]{2}{*}{argmax}  & none  
 & 0.73 & 0.62 & 0.63 & 0.67 & 0.78 & 0.69 & 0.69 & \textbf{0.72} \\
 &  & best                                                                
 & 0.72 & 0.59 & 0.65 & 0.66 & 0.77 & 0.71 & 0.68 & 0.71 \\
 & \multirow[t]{2}{*}{probs} & none                                         
 & 0.74 & 0.63 & 0.67 & 0.69 & 0.81 & \textbf{0.72} & \textbf{0.72} & 0.70 \\
 &  & best                                                                
 & 0.71 & 0.60 & 0.64 & 0.65 & 0.80 & 0.68 & 0.68 & 0.71 \\
\midrule
\multirow[t]{4}{*}{GPS} & \multirow[t]{2}{*}{argmax} & none 
 & 0.74 & 0.60 & 0.60 & 0.63 & 0.78 & 0.71 & 0.60 & 0.65 \\
 &  & best                                                
 & 0.77 & 0.62 & 0.67 & 0.67 & 0.76 & 0.68 & 0.66 & 0.71 \\
 & \multirow[t]{2}{*}{probs} & none                        
 & 0.74 & \textbf{0.64} & 0.61 & 0.66 & 0.80 & 0.69 & 0.60 & 0.66 \\
 &  & best                                                
 & 0.77 & 0.60 & 0.63 & 0.67 & 0.78 & 0.69 & 0.68 & 0.70 \\
\bottomrule

\end{tabular}
\end{table*}

\section{Properties of the Constructed Graphs}

The resulting graphs based on the original generated essays and their paraphrased variants contain between 8 and 10 edges per node in $T_1$, while Turkish backtranslation results in a range between 9 and 6 edges per node, and French in even fewer, 6 edges per node. With the higher thresholds, the number of edges per node drops consistently. The detailed dataset statistics are presented in Appendix \ref{app:graph_stats}.

Furthermore, in the original essays, there are very few isolated components: The proportion of graphs containing isolated components ranges between 0\% and 5\%, depending on the model family and version as well as the threshold. That proportion goes up to 8\% with paraphrasing and French backtranslation, while Gemma3 and Qwen3 essays backtranslated via Turkish contain the highest proportion of graphs with isolated components (15\% and 12\% respectively under $T_3$ threshold). 

The low proportion of graphs with isolated components is likely related to the fact that most of the graphs form a single source-sink path, meaning that these graphs have one starting node, one ending node and all of the intermediary nodes are connected in between -- these graphs make up over 90\% of all of the graphs for the original essays. Under the obfuscation strategies, this proportion decreases, however, it does not go below 70\%. 

Taken together, these observations indicate that the obtained reasoning graphs are dense and highly interconnected, with paraphrasing and backtranslation affecting the interconnectivity to varying degrees and suggesting that obfuscation strategies do not always perfectly preserve the reasoning structures of the original essays.




\section{Results}


\begin{table*}[t]
\small
\centering
\caption{Macro F1-score results in the \textsc{Cross-Version} setup. \textit{None} in Threshold column indicates a complete graph; \textit{best} refers to the best-performing threshold among $T_1, T_2, T_3$, which are defined in Section \ref{sec:experiments}. \textit{Argmax} and \textit{probs} refer to the edge processing strategies defined in Section \ref{sec:method}. `Orig.' refers to the non-obfuscated essays, `paraphr.' to the paraphrased, `BT-FR' and `BT-TR' to the backtranslated essays via French and Turkish, respectively.}
\label{tab:diffmodel-versions}

\begin{tabular}{lcccccccccc}
\toprule
& & &
\multicolumn{4}{c}{\makecell{LLM-OWL-AE-I train /\\ LLM-OWL-AE-II test}}
& \multicolumn{4}{c}{\makecell{LLM-OWL-AE-II train /\\ LLM-OWL-AE-I test}} \\
\cmidrule(lr){4-7} \cmidrule(lr){8-11}

Model & Edge & Thresh. 
& Orig. & Paraphr. & BT-FR & BT-TR
& Orig. & Paraphr. & BT-FR & BT-TR \\
\midrule

\multicolumn{11}{l}{\textit{Text-only baseline}} \\
Longformer         & -- & -- & 0.47 & 0.30 & 0.56 & 0.50 & 0.49 & 0.36 & 0.35 & 0.37 \\
\midrule

\multicolumn{11}{l}{\textit{Text + Reasoning Structure}} \\
\multirow[t]{4}{*}{GCN} & \multirow[t]{2}{*}{argmax} & none
 & 0.57 & 0.46 & 0.56 & 0.55 & 0.50 & 0.42 & 0.43 & 0.49 \\
 &  & best 
 & 0.60 & 0.48 & 0.56 & 0.57 & 0.49 & 0.41 & 0.42 & 0.46 \\
 & \multirow[t]{2}{*}{probs} & none
 & 0.57 & 0.46 & 0.55 & 0.55 & 0.50 & 0.42 & 0.43 & 0.49 \\
 &  & best 
 & 0.60 & 0.48 & 0.57 & 0.58 & 0.49 & 0.41 & 0.42 & 0.47 \\
\midrule
\multirow[t]{4}{*}{GAT} & \multirow[t]{2}{*}{argmax} & none
 & 0.62 & 0.50 & 0.61 & 0.61 & 0.51 & 0.43 & 0.42 & 0.50 \\
 &  & best 
 & 0.64 & 0.53 & 0.61 & 0.61 & \textbf{0.53} & 0.44 & 0.45 & 0.50 \\
 & \multirow[t]{2}{*}{probs} & none 
 & 0.63 & 0.50 & 0.60 & 0.59 & 0.51 & 0.43 & 0.43 & 0.49 \\
 &  & best 
 & 0.63 & 0.52 & 0.60 & 0.59 & \textbf{0.53} & 0.42 & 0.44 & 0.50 \\
\midrule
\multirow[t]{4}{*}{Graph Transformer} & \multirow[t]{2}{*}{argmax} & none
 & 0.65 & \textbf{0.54} & \textbf{0.64} & \textbf{0.63} & 0.52 & 0.45 & 0.44 & 0.49 \\
 &  & best 
 & \textbf{0.66} & 0.52 & 0.60 & 0.62 & 0.51 & 0.45 & 0.44 & 0.49 \\
 & \multirow[t]{2}{*}{probs} & none
 & 0.59 & 0.46 & 0.59 & 0.54 & 0.51 & \textbf{0.47} & 0.46 & 0.53 \\
 &  & best 
 & \textbf{0.66} & 0.53 & 0.61 & \textbf{0.63} & \textbf{0.53} & 0.42 & 0.44 & 0.49 \\
\midrule
\multirow[t]{4}{*}{GPS} & \multirow[t]{2}{*}{argmax} & none
 & 0.59 & 0.49 & 0.56 & 0.57 & \textbf{0.53} & 0.43 & 0.46 & 0.53 \\
 &  & best 
 & 0.64 & 0.53 & 0.62 & 0.61 & 0.50 & 0.43 & \textbf{0.48} & 0.55 \\
 & \multirow[t]{2}{*}{probs} & none
 & 0.59 & 0.49 & 0.55 & 0.58 & \textbf{0.53} & 0.46 & 0.46 & \textbf{0.57} \\
 &  & best 
 & 0.64 & 0.52 & 0.60 & 0.62 & \textbf{0.53} & 0.43 & 0.47 & 0.56 \\
\bottomrule

\end{tabular}
\end{table*}

\textbf{Reasoning graph-based approach generalises better than the text-only baseline when evaluated on the unseen model versions.} While Longformer achieves F1-scores exceeding 90\% in the \textsc{Same-Version} evaluation setup on the non-obfuscated essays as reported in Table \ref{tab:samemodel-versions}, the performance drops around or below 50\% in the \textsc{Cross-Version} setting (see Table \ref{tab:diffmodel-versions}). In comparison, a GNN model markedly outperforms Longformer by 19 percentage points when trained on the earlier model versions and evaluated on the later versions, with a more moderate improvement of 4 points when evaluated on the earlier model versions instead.
This might be explained by the fact that the later model versions incorporate and expand on the lexical representations generated by the earlier model versions, which would allow the text-only baseline to better generalise to the earlier model essays. 
However, reasoning features are more consistent across model versions, with less variation, allowing our reasoning graph-based approach to significantly outperform the baseline and show more stable results across the board.

\textbf{Reasoning graph-based approach exhibits more robust performance in most experimental configurations when compared to the text-only baseline.} In the \textsc{Same-Version} evaluation setup, reported in Table \ref{tab:samemodel-versions}, obfuscation attacks have a noticeable negative impact on the Longformer performance, with drops ranging 9 to 52 percentage points depending on the obfuscation strategy (with paraphrasing having the strongest effect) and model versions (stronger performance degradation is observed in later models). At the same time, while the GNN models perform worse than the baseline on the non-obfuscated data, they exhibit more consistent performance across obfuscation strategies, with the score drops ranging from 10 to 20 points. Furthermore, in the case of later model versions, they achieve up to 27 percentage point improvement over the baseline. In the \textsc{Cross-Version} evaluation setup, reported in Table \ref{tab:diffmodel-versions}, GNNs consistently outperform Longformer under obfuscation attacks with as much as a 24 percentage point improvement under paraphrasing when evaluated with the later model versions and 20 points under Turkish backtranslation when evaluated with the earlier model versions. 

\textbf{In terms of GNN architecture, Graph Transformer layer exhibits better performance across most experimental configurations.} Moreover, we find that fewer layers improve performance, with 1-layer configurations accounting for about 90\% of the best F1-scores across all GNN architectures. This is mainly due to the high connectivity of the generated graphs. Adding more layers means sharing more information across nodes and edges, which, with the structure of our reasoning graphs, ends up blurring many of the discriminative features. Finally, we could not observe a consistent optimal configuration in terms of edge thresholds and construction strategies. The best performance varies depending on the experiments and GNN architecture used.



\section{Conclusion}

With this work, we present the first study leveraging argumentative reasoning graphs for LLM-authorship attribution. We propose to go beyond surface-level lexical cues and instead incorporate reasoning information into authorship attribution. An existing argument mining pipeline allows us to construct reasoning graphs, which we then use to train and evaluate GNN models. Our extensive experimentation considers different authorship obfuscation techniques and evaluates across versions of models belonging to different families. We demonstrate that reasoning graph-based approach is more robust to obfuscation attacks than the standard fine-tuned Transformer-based approach, achieving up to a 27 percentage point increase, and exhibits better generalisation capabilities, indicated by up to a 19 percentage point increase when evaluated on the unseen model versions. 


\section*{Limitations}
We measure generalisation only in conditions in which the GNN model is trained and tested on the texts generated by the different versions of the LLM from the same family, in such a way approximating it across model versions. However, in future work, generalisation under different conditions, including different text genres and domains, should be considered. 

Another limitation concerns the reasoning graphs. The tasks of argument component detection and argument relation identification remain an open problem in the field of argument mining. Therefore, there is still room for improvement in terms of the quality of the constructed graphs, which could further strengthen the performance of our GNN-based approach.

\section*{Ethical Considerations}
Our work focuses on the task of LLM authorship attribution, and no human-authored data was produced or used during the experiments. As a result, there are no direct privacy concerns with respect to human authors. We acknowledge the risk of misuse of the authorship attribution systems for the purposes of surveilling LLM usage. Furthermore, the observations that were made regarding the impact of the obfuscation strategies and cross-model-version evaluation can inform evasion strategies. However, we believe that the benefits of such research outweigh the risks and highlight that our corpus and code are intended for research purposes only. All models used to generate data are open-source and accessed locally via Ollama. To the best of our knowledge, their respective licences place no restrictions on output generation for non-commercial research purposes as used in the present work.

\bibliography{custom}

@inproceedings{guo2024online,
  title={Online disinformation and generative language models: Motivations, challenges, and mitigations},
  author={Guo, Ziyi},
  booktitle={Companion Proceedings of the ACM Web Conference 2024},
  pages={1174--1177},
  year={2024}
}

@article{krishna2023paraphrasing,
  title={Paraphrasing evades detectors of ai-generated text, but retrieval is an effective defense},
  author={Krishna, Kalpesh and Song, Yixiao and Karpinska, Marzena and Wieting, John and Iyyer, Mohit},
  journal={Advances in neural information processing systems},
  volume={36},
  pages={27469--27500},
  year={2023}
}

@article{kasneci2023chatgpt,
  title={ChatGPT for good? On opportunities and challenges of large language models for education},
  author={Kasneci, Enkelejda and Se{\ss}ler, Kathrin and K{\"u}chemann, Stefan and Bannert, Maria and Dementieva, Daryna and Fischer, Frank and Gasser, Urs and Groh, Georg and G{\"u}nnemann, Stephan and H{\"u}llermeier, Eyke and others},
  journal={Learning and individual differences},
  volume={103},
  pages={102274},
  year={2023},
  publisher={Elsevier}
}

@article{wu2025survey,
  title={A survey on llm-generated text detection: Necessity, methods, and future directions},
  author={Wu, Junchao and Yang, Shu and Zhan, Runzhe and Yuan, Yulin and Chao, Lidia Sam and Wong, Derek Fai},
  journal={Computational Linguistics},
  volume={51},
  number={1},
  pages={275--338},
  year={2025}
}

@article{liu2023argugpt,
  title={ArguGPT: evaluating, understanding and identifying argumentative essays generated by GPT models},
  author={Liu, Yikang and Zhang, Ziyin and Zhang, Wanyang and Yue, Shisen and Zhao, Xiaojing and Cheng, Xinyuan and Zhang, Yiwen and Hu, Hai},
  journal={arXiv preprint arXiv:2304.07666},
  year={2023}
}

@inproceedings{abassy2024llm,
    title = "{LLM}-{D}etect{AI}ve: a Tool for Fine-Grained Machine-Generated Text Detection",
    author = "Abassy, Mervat  and
      Elozeiri, Kareem  and
      Aziz, Alexander  and
      Ta, Minh Ngoc  and
      Tomar, Raj Vardhan  and
      Adhikari, Bimarsha  and
      Ahmed, Saad El Dine  and
      Wang, Yuxia  and
      Mohammed Afzal, Osama  and
      Xie, Zhuohan  and
      Mansurov, Jonibek  and
      Artemova, Ekaterina  and
      Mikhailov, Vladislav  and
      Xing, Rui  and
      Geng, Jiahui  and
      Iqbal, Hasan  and
      Mujahid, Zain Muhammad  and
      Mahmoud, Tarek  and
      Tsvigun, Akim  and
      Aji, Alham Fikri  and
      Shelmanov, Artem  and
      Habash, Nizar  and
      Gurevych, Iryna  and
      Nakov, Preslav",
    booktitle = "Proceedings of the 2024 Conference on Empirical Methods in Natural Language Processing: System Demonstrations",
    year = "2024",
    url = "https://aclanthology.org/2024.emnlp-demo.35/",
    pages = "336--343"
}

@inproceedings{dugan2024raid,
  title={Raid: A shared benchmark for robust evaluation of machine-generated text detectors},
  author={Dugan, Liam and Hwang, Alyssa and Trhl{\'\i}k, Filip and Zhu, Andrew and Ludan, Josh Magnus and Xu, Hainiu and Ippolito, Daphne and Callison-Burch, Chris},
  booktitle={Proceedings of the 62nd Annual Meeting of the Association for Computational Linguistics (Volume 1: Long Papers)},
  pages={12463--12492},
  year={2024}
}

@inproceedings{yu2025evobench,
  title={Evobench: Towards real-world llm-generated text detection benchmarking for evolving large language models},
  author={Yu, Xiao and Yu, Yi and Liu, Dongrui and Chen, Kejiang and Zhang, Weiming and Yu, Nenghai and Shao, Jing},
  booktitle={Findings of the Association for Computational Linguistics: ACL 2025},
  pages={14605--14620},
  year={2025}
}

@inproceedings{kumarage2023neural,
  title={Neural authorship attribution: Stylometric analysis on large language models},
  author={Kumarage, Tharindu and Liu, Huan},
  booktitle={2023 International conference on cyber-enabled distributed computing and knowledge discovery (cyberc)},
  pages={51--54},
  year={2023},
  organization={IEEE}
}

@article{huang2025authorship,
  title={Authorship attribution in the era of llms: Problems, methodologies, and challenges},
  author={Huang, Baixiang and Chen, Canyu and Shu, Kai},
  journal={ACM SIGKDD Explorations Newsletter},
  volume={26},
  number={2},
  pages={21--43},
  year={2025},
  publisher={ACM New York, NY, USA}
}

@inproceedings{bevendorff2025two,
  title={The Two Paradigms of LLM Detection: Authorship Attribution vs Authorship Verification},
  author={Bevendorff, Janek and Wiegmann, Matti and Richter, Emmelie and Potthast, Martin and Stein, Benno},
  booktitle={Findings of the Association for Computational Linguistics: ACL 2025},
  pages={3762--3787},
  year={2025}
}

@article{li2023origin,
  title={Origin tracing and detecting of llms},
  author={Li, Linyang and Wang, Pengyu and Ren, Ke and Sun, Tianxiang and Qiu, Xipeng},
  journal={arXiv preprint arXiv:2304.14072},
  year={2023}
}

@inproceedings{yang2023dna,
  title={Dna-gpt: Divergent n-gram analysis for training-free detection of gpt-generated text},
  author={Yang, Xianjun and Cheng, Wei and Wu, Yue and Petzold, Linda Ruth and Wang, William Yang and Chen, Haifeng},
  booktitle={The Twelfth International Conference on Learning Representations},
  year={2023}
}

@inproceedings{tiedemann-thottingal-2020-opus,
    title = "{OPUS}-{MT} {--} Building open translation services for the World",
    author = {Tiedemann, J{\"o}rg  and Thottingal, Santhosh},
    booktitle = "Proceedings of the 22nd Annual Conference of the European Association for Machine Translation",
    month = nov,
    year = "2020",
    address = "Lisboa, Portugal",
    publisher = "European Association for Machine Translation",
    url = "https://aclanthology.org/2020.eamt-1.61",
    pages = "479--480",
}

@inproceedings{tiedemann-2020-tatoeba,
    title = "The Tatoeba Translation Challenge {--} Realistic Data Sets for Low Resource and Multilingual {MT}",
    author = {Tiedemann, J{\"o}rg},
    booktitle = "Proceedings of the Fifth Conference on Machine Translation",
    month = nov,
    year = "2020",
    address = "Online",
    publisher = "Association for Computational Linguistics",
    url = "https://aclanthology.org/2020.wmt-1.139",
    pages = "1174--1182",
}

@inproceedings{macko2024authorship,
  title={Authorship obfuscation in multilingual machine-generated text detection},
  author={Macko, Dominik and Moro, Robert and Uchendu, Adaku and Srba, Ivan and Lucas, Jason S and Yamashita, Michiharu and Tripto, Nafis Irtiza and Lee, Dongwon and Simko, Jakub and Bielikova, Maria},
  booktitle={Findings of the Association for Computational Linguistics: EMNLP 2024},
  pages={6348--6368},
  year={2024}
}

@inproceedings{altakrori-etal-2022-multifaceted,
    title = "A Multifaceted Framework to Evaluate Evasion, Content Preservation, and Misattribution in Authorship Obfuscation Techniques",
    author = "Altakrori, Malik  and
      Scialom, Thomas  and
      Fung, Benjamin C. M.  and
      Cheung, Jackie Chi Kit",
    editor = "Goldberg, Yoav  and
      Kozareva, Zornitsa  and
      Zhang, Yue",
    booktitle = "Proceedings of the 2022 Conference on Empirical Methods in Natural Language Processing",
    month = dec,
    year = "2022",
    address = "Abu Dhabi, United Arab Emirates",
    publisher = "Association for Computational Linguistics",
    url = "https://aclanthology.org/2022.emnlp-main.153/",
    doi = "10.18653/v1/2022.emnlp-main.153",
    pages = "2391--2406",
}

@inproceedings{ayoobi2025esperanto,
  title={Esperanto: Evaluating synthesized phrases to enhance robustness in ai detection for text origination},
  author={Ayoobi, Navid and Knab, Lily and Cheng, Wen and Pantoja, David and Alikhani, Hamidreza and Flamant, Sylvain and Kim, Jin and Mukherjee, Arjun},
  booktitle={Proceedings of the 36th ACM Conference on Hypertext and Social Media},
  pages={1--10},
  year={2025}
}

@inproceedings{li-etal-2025-prdetect,
    title = "{PRD}etect: Perturbation-Robust {LLM}-generated Text Detection Based on Syntax Tree",
    author = "Li, Xiang  and
      Yin, Zhiyi  and
      Tan, Hexiang  and
      Jing, Shaoling  and
      Su, Du  and
      Cheng, Yi  and
      Shen, Huawei  and
      Sun, Fei",
    editor = "Chiruzzo, Luis  and
      Ritter, Alan  and
      Wang, Lu",
    booktitle = "Findings of the Association for Computational Linguistics: NAACL 2025",
    month = apr,
    year = "2025",
    address = "Albuquerque, New Mexico",
    publisher = "Association for Computational Linguistics",
    url = "https://aclanthology.org/2025.findings-naacl.464/",
    doi = "10.18653/v1/2025.findings-naacl.464",
    pages = "8305--8316",
    ISBN = "979-8-89176-195-7",
}

@article{ruiz2024nlas,
  title={Nlas-multi: A multilingual corpus of automatically generated natural language argumentation schemes},
  author={Ruiz-Dolz, Ramon and Taverner, Joaquin and Lawrence, John and Reed, Chris},
  journal={Data in Brief},
  volume={57},
  pages={111087},
  year={2024},
  publisher={Elsevier}
}

@article{beltagy2020longformer,
  title={Longformer: The long-document transformer},
  author={Beltagy, Iz and Peters, Matthew E and Cohan, Arman},
  journal={arXiv preprint arXiv:2004.05150},
  year={2020}
}

@inproceedings{wang2024m4gt,
  title={M4gt-bench: Evaluation benchmark for black-box machine-generated text detection},
  author={Wang, Yuxia and Mansurov, Jonibek and Ivanov, Petar and Su, Jinyan and Shelmanov, Artem and Tsvigun, Akim and Afzal, Osama Mohammed and Mahmoud, Tarek and Puccetti, Giovanni and Arnold, Thomas and others},
  booktitle={Proceedings of the 62nd Annual Meeting of the Association for Computational Linguistics (Volume 1: Long Papers)},
  pages={3964--3992},
  year={2024}
}

@inproceedings{macko2023multitude,
  title={MULTITuDE: Large-scale multilingual machine-generated text detection benchmark},
  author={Macko, Dominik and Moro, Robert and Uchendu, Adaku and Lucas, Jason and Yamashita, Michiharu and Pikuliak, Mat{\'u}{\v{s}} and Srba, Ivan and Le, Thai and Lee, Dongwon and Simko, Jakub and others},
  booktitle={Proceedings of the 2023 Conference on Empirical Methods in Natural Language Processing},
  pages={9960--9987},
  year={2023}
}

@inproceedings{la2025openturingbench,
  title={OpenTuringBench: An Open-Model-based Benchmark and Framework for Machine-Generated Text Detection and Attribution},
  author={La Cava, Lucio and Tagarelli, Andrea},
  booktitle={Proceedings of the 2025 Conference on Empirical Methods in Natural Language Processing},
  pages={26666--26682},
  year={2025}
}

@inproceedings{zhou2024humanizing,
  title={Humanizing machine-generated content: evading AI-text detection through adversarial attack},
  author={Zhou, Ying and He, Ben and Sun, Le},
  booktitle={Proceedings of the 2024 Joint International Conference on Computational Linguistics, Language Resources and Evaluation (LREC-COLING 2024)},
  pages={8427--8437},
  year={2024}
}

@article{uchendu2023attribution,
  title={Attribution and obfuscation of neural text authorship: A data mining perspective},
  author={Uchendu, Adaku and Le, Thai and Lee, Dongwon},
  journal={ACM SIGKDD Explorations Newsletter},
  volume={25},
  number={1},
  pages={1--18},
  year={2023},
  publisher={ACM New York, NY, USA}
}

@article{sadasivan2023can,
  title={Can AI-generated text be reliably detected?},
  author={Sadasivan, Vinu Sankar and Kumar, Aounon and Balasubramanian, Sriram and Wang, Wenxiao and Feizi, Soheil},
  journal={arXiv preprint arXiv:2303.11156},
  year={2023}
}

@inproceedings{wang2024raft,
  title={RAFT: Realistic attacks to fool text detectors},
  author={Wang, James Liyuan and Li, Ran and Yang, Junfeng and Mao, Chengzhi},
  booktitle={Proceedings of the 2024 Conference on Empirical Methods in Natural Language Processing},
  pages={16923--16936},
  year={2024}
}

@article{shi2024red,
  title={Red teaming language model detectors with language models},
  author={Shi, Zhouxing and Wang, Yihan and Yin, Fan and Chen, Xiangning and Chang, Kai-Wei and Hsieh, Cho-Jui},
  journal={Transactions of the Association for Computational Linguistics},
  volume={12},
  pages={174--189},
  year={2024},
  publisher={MIT Press One Broadway, 12th Floor, Cambridge, Massachusetts 02142, USA~…}
}

@inproceedings{fang2025your,
  title={Your language model can secretly write like humans: Contrastive paraphrase attacks on llm-generated text detectors},
  author={Fang, Hao and Kong, Jiawei and Zhuang, Tianqu and Qiu, Yixiang and Gao, Kuofeng and Chen, Bin and Xia, Shu-Tao and Wang, Yaowei and Zhang, Min},
  booktitle={Proceedings of the 2025 Conference on Empirical Methods in Natural Language Processing},
  pages={8596--8613},
  year={2025}
}

@inproceedings{zhou2024navigating,
  title={Navigating the shadows: Unveiling effective disturbances for modern ai content detectors},
  author={Zhou, Ying and He, Ben and Sun, Le},
  booktitle={Proceedings of the 62nd Annual Meeting of the Association for Computational Linguistics (Volume 1: Long Papers)},
  pages={10847--10861},
  year={2024}
}

@inproceedings{chakraborty-etal-2023-counter,
    title = "Counter {T}uring Test ({CT}2): {AI}-Generated Text Detection is Not as Easy as You May Think - Introducing {AI} Detectability Index ({ADI})",
    author = "Chakraborty, Megha  and
      Tonmoy, S.M Towhidul Islam  and
      Zaman, S M Mehedi  and
      Gautam, Shreya  and
      Kumar, Tanay  and
      Sharma, Krish  and
      Barman, Niyar  and
      Gupta, Chandan  and
      Jain, Vinija  and
      Chadha, Aman  and
      Sheth, Amit  and
      Das, Amitava",
    editor = "Bouamor, Houda  and
      Pino, Juan  and
      Bali, Kalika",
    booktitle = "Proceedings of the 2023 Conference on Empirical Methods in Natural Language Processing",
    month = dec,
    year = "2023",
    address = "Singapore",
    publisher = "Association for Computational Linguistics",
    url = "https://aclanthology.org/2023.emnlp-main.136/",
    doi = "10.18653/v1/2023.emnlp-main.136",
    pages = "2206--2239",
}

@inproceedings{gemechu-etal-2025-open,
    title = "The Open Argument Mining Framework",
    author = "Gemechu, Debela  and
      Ruiz-Dolz, Ramon  and
      G{\'o}rska, Kamila  and
      Moslemnejad, Somaye  and
      Maguire, Eimear  and
      Zografistou, Dimitra  and
      Jo, Yohan  and
      Lawrence, John  and
      Reed, Chris",
    editor = "Mishra, Pushkar  and
      Muresan, Smaranda  and
      Yu, Tao",
    booktitle = "Proceedings of the 63rd Annual Meeting of the Association for Computational Linguistics (Volume 3: System Demonstrations)",
    month = jul,
    year = "2025",
    address = "Vienna, Austria",
    publisher = "Association for Computational Linguistics",
    url = "https://aclanthology.org/2025.acl-demo.31/",
    doi = "10.18653/v1/2025.acl-demo.31",
    pages = "318--328",
}

@inproceedings{chernodub-etal-2019-targer,
    title = "{TARGER}: Neural Argument Mining at Your Fingertips",
    author = "Chernodub, Artem  and
      Oliynyk, Oleksiy  and
      Heidenreich, Philipp  and
      Bondarenko, Alexander  and
      Hagen, Matthias  and
      Biemann, Chris  and
      Panchenko, Alexander",
    editor = "Costa-juss{\`a}, Marta R.  and
      Alfonseca, Enrique",
    booktitle = "Proceedings of the 57th Annual Meeting of the Association for Computational Linguistics: System Demonstrations",
    month = jul,
    year = "2019",
    address = "Florence, Italy",
    publisher = "Association for Computational Linguistics",
    url = "https://aclanthology.org/P19-3031/",
    doi = "10.18653/v1/P19-3031",
    pages = "195--200",
}

@article{ruiz2021transformer,
  title={Transformer-based models for automatic identification of argument relations: A cross-domain evaluation},
  author={Ruiz-Dolz, Ramon and Alemany, Jose and Barber{\'a}, Stella M Heras and Garc{\'\i}a-Fornes, Ana},
  journal={IEEE Intelligent Systems},
  volume={36},
  number={6},
  pages={62--70},
  year={2021},
  publisher={IEEE}
}

@inproceedings{reimers-2019-sentence-bert,
  title = "Sentence-BERT: Sentence Embeddings using Siamese BERT-Networks",
  author = "Reimers, Nils and Gurevych, Iryna",
  booktitle = "Proceedings of the 2019 Conference on Empirical Methods in Natural Language Processing",
  month = "11",
  year = "2019",
  publisher = "Association for Computational Linguistics",
  url = "https://arxiv.org/abs/1908.10084",
}

@inproceedings{cibier2024graph,
  title={Graph Convolutional Networks and Graph Attention Networks for Approximating Arguments Acceptability},
  author={Cibier, Paul and Mailly, Jean-Guy},
  booktitle={10th International Conference on Computational Models of Argument (COMMA 2024)},
  year={2024},
  organization={IOS Press}
}

@inproceedings{kuhlmann2019using,
  title={Using graph convolutional networks for approximate reasoning with abstract argumentation frameworks: A feasibility study},
  author={Kuhlmann, Isabelle and Thimm, Matthias},
  booktitle={International Conference on Scalable Uncertainty Management},
  pages={24--37},
  year={2019},
  organization={Springer}
}

@inproceedings{gehlot2026heterogeneous,
  title={Heterogeneous Graph Neural Networks for Assumption-Based Argumentation},
  author={Gehlot, Preesha and Rapberger, Anna and Russo, Fabrizio and Toni, Francesca},
  booktitle={Proceedings of the AAAI Conference on Artificial Intelligence},
  volume={40},
  number={23},
  pages={19117--19125},
  year={2026}
}

@inproceedings{malmqvist2020determining,
  title={Determining the Acceptability of Abstract Arguments with Graph Convolutional Networks.},
  author={Malmqvist, Lars and Yuan, Tommy and Nightingale, Peter and Manandhar, Suresh},
  booktitle={SAFA@ COMMA},
  pages={47--56},
  year={2020}
}

@article{ruggeri2021tree,
  title={Tree-constrained graph neural networks for argument mining},
  author={Ruggeri, Federico and Lippi, Marco and Torroni, Paolo},
  journal={arXiv preprint arXiv:2110.00124},
  year={2021}
}

@inproceedings{ruiz2023automatic,
  title={Automatic debate evaluation with argumentation semantics and natural language argument graph networks},
  author={Ruiz-Dolz, Ramon and Heras, Stella and Garcia, Ana},
  booktitle={Proceedings of the 2023 conference on empirical methods in natural language processing},
  pages={6030--6040},
  year={2023}
}

@inproceedings{Zhang-arg,
author = {Zhang, Gechuan and Nulty, Paul and Lillis, David},
title = {Argument Mining with Graph Representation Learning},
year = {2023},
isbn = {9798400701979},
publisher = {Association for Computing Machinery},
address = {New York, NY, USA},
url = {https://doi.org/10.1145/3594536.3595152},
doi = {10.1145/3594536.3595152},
booktitle = {Proceedings of the Nineteenth International Conference on Artificial Intelligence and Law},
pages = {371–380},
numpages = {10},
location = {Braga, Portugal},
series = {ICAIL '23}
}

@article{mao2024seeing,
  title={Seeing both sides: context-aware heterogeneous graph matching networks for extracting-related arguments},
  author={Mao, Tiezheng and Yoshie, Osamu and Fu, Jialing and Mao, Weixin},
  journal={Neural Computing and Applications},
  volume={36},
  number={9},
  pages={4741--4762},
  year={2024},
  publisher={Springer}
}

@inproceedings{Kipf-GCN,
  author       = {Thomas N. Kipf and
                  Max Welling},
  title        = {Semi-Supervised Classification with Graph Convolutional Networks},
  booktitle    = {5th International Conference on Learning Representations, {ICLR} 2017,
                  Toulon, France, April 24-26, 2017, Conference Track Proceedings},
  publisher    = {OpenReview.net},
  year         = {2017},
  url          = {https://openreview.net/forum?id=SJU4ayYgl},
  bibsource    = {dblp computer science bibliography, https://dblp.org}
}

@article{velikovi2017graph,
  author = {Veličković, Petar and Cucurull, Guillem and Casanova, Arantxa and Romero, Adriana and Liò, Pietro and Bengio, Yoshua},
  journal = {6th International Conference on Learning Representations},
  title = {Graph Attention Networks},
  year = 2017
}

@inproceedings{shi2021masked,
  title={Masked Label Prediction: Unified Message Passing Model for Semi-Supervised Classification},
  author={Shi, Yunsheng and Huang, Zhengjie and Feng, Shikun and Zhong, Hui and Wang, Wenjing and Sun, Yu},
  booktitle={Proceedings of the Thirtieth International Joint Conference on Artificial Intelligence},
  pages={1548--1554},
  year={2021},
  organization={International Joint Conferences on Artificial Intelligence Organization}
}

@article{rampavsek2022recipe,
  title={Recipe for a general, powerful, scalable graph transformer},
  author={Ramp{\'a}{\v{s}}ek, Ladislav and Galkin, Michael and Dwivedi, Vijay Prakash and Luu, Anh Tuan and Wolf, Guy and Beaini, Dominique},
  journal={Advances in Neural Information Processing Systems},
  volume={35},
  pages={14501--14515},
  year={2022}
}

@article{mccarthy2010mtld,
  title={{MTLD, vocd-D, and HD-D: A validation study of sophisticated approaches to lexical diversity assessment}},
  author={McCarthy, Philip M and Jarvis, Scott},
  journal={Behavior Research Methods},
  volume={42},
  number={2},
  pages={381--392},
  year={2010},
  publisher={Springer}
}

@inproceedings{sun-etal-2025-ai,
    title = "Are We in the {AI}-Generated Text World Already? Quantifying and Monitoring {AIGT} on Social Media",
    author = "Sun, Zhen  and
      Zhang, Zongmin  and
      Shen, Xinyue  and
      Zhang, Ziyi  and
      Liu, Yule  and
      Backes, Michael  and
      Zhang, Yang  and
      He, Xinlei",
    editor = "Che, Wanxiang  and
      Nabende, Joyce  and
      Shutova, Ekaterina  and
      Pilehvar, Mohammad Taher",
    booktitle = "Proceedings of the 63rd Annual Meeting of the Association for Computational Linguistics (Volume 1: Long Papers)",
    month = jul,
    year = "2025",
    address = "Vienna, Austria",
    publisher = "Association for Computational Linguistics",
    url = "https://aclanthology.org/2025.acl-long.1120/",
    doi = "10.18653/v1/2025.acl-long.1120",
    pages = "22975--23005",
    ISBN = "979-8-89176-251-0"
}

@article{dolezal2026impact,
  title={The Impact of AI-Generated Text on the Internet},
  author={Dolezal, Jonas and Alam, Sawood and Graham, Mark and Bohacek, Maty},
  journal={arXiv preprint arXiv:2604.26965},
  year={2026}
}

@inproceedings{vykopal-etal-2024-disinformation,
    title = "Disinformation Capabilities of Large Language Models",
    author = "Vykopal, Ivan  and
      Pikuliak, Mat{\'u}{\v{s}}  and
      Srba, Ivan  and
      Moro, Robert  and
      Macko, Dominik  and
      Bielikova, Maria",
    editor = "Ku, Lun-Wei  and
      Martins, Andre  and
      Srikumar, Vivek",
    booktitle = "Proceedings of the 62nd Annual Meeting of the Association for Computational Linguistics (Volume 1: Long Papers)",
    month = aug,
    year = "2024",
    address = "Bangkok, Thailand",
    publisher = "Association for Computational Linguistics",
    url = "https://aclanthology.org/2024.acl-long.793/",
    doi = "10.18653/v1/2024.acl-long.793",
    pages = "14830--14847"
}

@article{buyl2026large,
  title={Large language models reflect the ideology of their creators},
  author={Buyl, Maarten and Rogiers, Alexander and Noels, Sander and Bied, Guillaume and Dominguez-Catena, Iris and Heiter, Edith and Johary, Iman and Mara, Alexandru-Cristian and Romero, Rapha{\"e}l and Lijffijt, Jefrey and others},
  journal={npj Artificial Intelligence},
  volume={2},
  number={1},
  pages={7},
  year={2026},
  publisher={Nature Publishing Group UK London}
}

@inproceedings{bang2024measuring,
  title={Measuring political bias in large language models: What is said and how it is said},
  author={Bang, Yejin and Chen, Delong and Lee, Nayeon and Fung, Pascale},
  booktitle={Proceedings of the 62nd Annual Meeting of the Association for Computational Linguistics (Volume 1: Long Papers)},
  pages={11142--11159},
  year={2024}
}

@inproceedings{mitchell2023detectgpt,
  title={Detectgpt: Zero-shot machine-generated text detection using probability curvature},
  author={Mitchell, Eric and Lee, Yoonho and Khazatsky, Alexander and Manning, Christopher D and Finn, Chelsea},
  booktitle={International conference on machine learning},
  pages={24950--24962},
  year={2023},
  organization={PMLR}
}

@inproceedings{hans2024spotting,
  title={Spotting LLMs with binoculars: zero-shot detection of machine-generated text},
  author={Hans, Abhimanyu and Schwarzschild, Avi and Cherepanova, Valeriia and Kazemi, Hamid and Saha, Aniruddha and Goldblum, Micah and Geiping, Jonas and Goldstein, Tom},
  booktitle={Proceedings of the 41st International Conference on Machine Learning},
  pages={17519--17537},
  year={2024}
}

@article{Li_Han_Wu_2018, title={Deeper Insights Into Graph Convolutional Networks for Semi-Supervised Learning}, volume={32}, url={https://ojs.aaai.org/index.php/AAAI/article/view/11604}, DOI={10.1609/aaai.v32i1.11604}, abstractNote={ &amp;lt;p&amp;gt; &amp;lt;div class=&amp;quot;page&amp;quot; title=&amp;quot;Page 1&amp;quot;&amp;gt;&amp;lt;div class=&amp;quot;layoutArea&amp;quot;&amp;gt;&amp;lt;div class=&amp;quot;column&amp;quot;&amp;gt;&amp;lt;p&amp;gt;&amp;lt;span&amp;gt;Many interesting problems in machine learning are being revisited with new deep learning tools. For graph-based semi-supervised learning, a recent important development is graph convolutional networks (GCNs), which nicely integrate local vertex features and graph topology in the convolutional layers. Although the GCN model compares favorably with other state-of-the-art methods, its mechanisms are not clear and it still requires considerable amount of labeled data for validation and model selection. &amp;lt;/span&amp;gt;In this paper, we develop deeper insights into the GCN model and address its fundamental limits. First, we show that the graph convolution of the GCN model is actually a special form of Laplacian smoothing, which is the key reason why GCNs work, but it also brings potential concerns of over-smoothing with many convolutional layers. Second, to overcome the limits of the GCN model with shallow architectures, we propose both co-training and self-training approaches to train GCNs. Our approaches significantly improve GCNs in learning with very few labels, and exempt them from requiring additional labels for validation. Extensive experiments on benchmarks have verified our theory and proposals.&amp;lt;/p&amp;gt;&amp;lt;/div&amp;gt;&amp;lt;/div&amp;gt;&amp;lt;/div&amp;gt; &amp;lt;/p&amp;gt; }, number={1}, journal={Proceedings of the AAAI Conference on Artificial Intelligence}, author={Li, Qimai and Han, Zhichao and Wu, Xiao-ming}, year={2018}, month={Apr.} }

@misc{lexrich,
        author = {Shen, Lucas},
        doi = {10.5281/zenodo.6607007},
        license = {MIT license},
        title = {{LexicalRichness: A small module to compute textual lexical richness}},
        url = {https://github.com/LSYS/lexicalrichness},
        year = {2022}
}

@article{Honnibal_spaCy,
author = {Honnibal, Matthew and Montani, Ines and Van Landeghem, Sofie and Boyd, Adriane},
doi = {10.5281/zenodo.1212303},
title = {{spaCy: Industrial-strength Natural Language Processing in Python}},
year = {2020}
}

@inproceedings{xing2024alison,
  title={Alison: Fast and effective stylometric authorship obfuscation},
  author={Xing, Eric and Venkatraman, Saranya and Le, Thai and Lee, Dongwon},
  booktitle={Proceedings of the AAAI Conference on Artificial Intelligence},
  volume={38},
  number={17},
  pages={19315--19322},
  year={2024}
}

@article{elkins2020can,
  title={Can GPT-3 pass a writer’s Turing test?},
  author={Elkins, Katherine and Chun, Jon},
  journal={Journal of Cultural Analytics},
  volume={5},
  number={2},
  year={2020}
}

@article{gao2022comparing,
  title={Comparing scientific abstracts generated by ChatGPT to original abstracts using an artificial intelligence output detector, plagiarism detector, and blinded human reviewers},
  author={Gao, Catherine A and Howard, Frederick M and Markov, Nikolay S and Dyer, Emma C and Ramesh, Siddhi and Luo, Yuan and Pearson, Alexander T},
  journal={BioRxiv},
  year={2022},
  publisher={Cold Spring Harbor Laboratory}
}

@article{cotton2023chatting,
  author = {Cotton, Debby R. E. and Cotton, Peter A. and Shipway, J. Reuben},
  title = {Chatting and Cheating: Ensuring Academic Integrity in the Era of {ChatGPT}},
  year = {2023},
  journal = {EdArXiv},
}

@article{dehouche2021plagiarism,
  title={Plagiarism in the age of massive Generative Pre-trained Transformers (GPT-3)},
  author={Dehouche, Nassim},
  journal={Ethics in Science and Environmental Politics},
  volume={21},
  pages={17--23},
  year={2021}
}
\appendix
\cleardoublepage
\onecolumn
\section{Linguistic Feature Statistics}
\label{app:ling_stats}
We use LexicalRichness Python module to calculate essay MTLD scores \cite{lexrich}. We use spaCy (model \texttt{en\_core\_web\_sm}) \cite{Honnibal_spaCy} to extract sentence clauses (\texttt{acl}, \texttt{conj}, \texttt{advcl}, \texttt{ccomp}, \texttt{csubj}, \texttt{discourse}, \texttt{parataxis}). The number of extracted clauses per sentence is then used to approximate syntactic complexity.

\begin{table*}[h!]
\small
\centering
\caption{Essays Linguistic Features (number of words, MTLD score and measure of syntactic complexity) in the LLM-OWL-AE-I data partition. We report mean and standard deviation.}
\label{tab:lex-features}

\begin{tabular}{llcccc}
\toprule
&
& \multicolumn{4}{c}{LLM-OWL-AE-I} \\
\cmidrule(lr){3-6}

Obfus. & Feature
& Gemma3 & Qwen3 & Llama3 & Phi3 \\
\midrule

\multirow{3}{*}{Orig.} & Words & 543 $\pm$ 26 & 481 $\pm$ 40 & 532 $\pm$ 39 & 504 $\pm$ 55  \\
 & MTLD & 160 $\pm$ 28 & 184 $\pm$ 31 & 103 $\pm$ 17 & 180 $\pm$ 25  \\
 & Syntactic compl. & 2.33 $\pm$ 0.28 & 2.00 $\pm$ 0.28 & 2.44 $\pm$ 0.43 & 2.22 $\pm$ 0.34 \\
\midrule
\multirow{3}{*}{Paraphr.} & Words & 557 $\pm$ 42 & 508 $\pm$ 54 & 495 $\pm$ 46 & 483 $\pm$ 61 \\
 & MTLD & 85 $\pm$ 18 & 96 $\pm$ 20 & 78 $\pm$ 13 & 94 $\pm$ 18 \\
 & Syntactic compl. & 2.17 $\pm$ 0.34 & 1.96 $\pm$ 0.35 & 2.30 $\pm$ 0.41 & 2.06 $\pm$ 0.36  \\
\midrule
\multirow{3}{*}{BT-FR} & Words & 367 $\pm$ 40 & 327 $\pm$ 40 & 375 $\pm$ 48 & 391 $\pm$ 57 \\
 & MTLD & 127 $\pm$ 26 & 142 $\pm$ 27 & 92 $\pm$ 17 & 135 $\pm$ 22 \\
 & Syntactic compl. & 2.57 $\pm$ 0.61 & 2.27 $\pm$ 0.52 & 2.93 $\pm$ 1.06 & 2.58 $\pm$ 0.65 \\
\midrule
\multirow{3}{*}{BT-TR} & Words & 392 $\pm$ 31 & 384 $\pm$ 33 & 437 $\pm$ 42 & 481 $\pm$ 56  \\
 & MTLD & 134 $\pm$ 25 & 147 $\pm$ 26 & 94 $\pm$ 16 & 148 $\pm$ 23 \\
 & Syntactic compl. & 2.04 $\pm$ 0.31 & 1.85 $\pm$ 0.31 & 2.28 $\pm$ 0.41 & 2.18 $\pm$ 0.38  \\
 \bottomrule
\end{tabular}
\end{table*}

\begin{table*}[h!]
\small
\centering
\caption{Essay linguistic Features (number of words, MTLD score and measure of syntactic complexity) in the LLM-OWL-AE-II data partition. We report mean and standard deviation.}
\label{tab:lex-features}

\begin{tabular}{llcccccccc}
\toprule
&
& \multicolumn{4}{c}{LLM-OWL-AE-II} \\
\cmidrule(lr){3-6} 

Obfus. & Feature
& Gemma4 & Qwen3.5 & Llama4 & Phi4 \\
\midrule

\multirow{3}{*}{Orig.} & Words & 466 $\pm$ 20 & 476 $\pm$ 34 & 477 $\pm$ 37 & 487 $\pm$ 39 \\
 & MTLD & 129 $\pm$ 26 & 185 $\pm$ 37 & 88 $\pm$ 15 & 196 $\pm$ 28 \\
 & Syntactic compl. & 1.91 $\pm$ 0.28 & 1.85 $\pm$ 0.33 & 2.11 $\pm$ 0.39 & 2.07 $\pm$ 0.30 \\
\midrule
\multirow{3}{*}{Paraphr.} & Words & 465 $\pm$ 36 & 499 $\pm$ 46 & 447 $\pm$ 45 & 494 $\pm$ 50 \\
 & MTLD & 79 $\pm$ 17 & 90 $\pm$ 21 & 72 $\pm$ 13 & 94 $\pm$ 19 \\
 & Syntactic compl. & 1.88 $\pm$ 0.34 & 1.78 $\pm$ 0.33 & 2.04 $\pm$ 0.41 & 1.99 $\pm$ 0.38 \\
\midrule
\multirow{3}{*}{BT-FR} & Words & 349 $\pm$ 37 & 322 $\pm$ 41 & 358 $\pm$ 44 & 356 $\pm$ 41 \\
 & MTLD & 113 $\pm$ 26 & 146 $\pm$ 32 & 83 $\pm$ 15 & 143 $\pm$ 24 \\
 & Syntactic compl. & 2.01 $\pm$ 0.46 & 1.95 $\pm$ 0.63 & 2.58 $\pm$ 0.77 & 2.46 $\pm$ 0.63 \\
\midrule
\multirow{3}{*}{BT-TR} & Words & 388 $\pm$ 24 & 345 $\pm$ 40 & 438 $\pm$ 35 & 456 $\pm$ 41 \\
 & MTLD & 120 $\pm$ 24 & 152 $\pm$ 33 & 84 $\pm$ 14 & 155 $\pm$ 24 \\
 & Syntactic compl. & 1.74 $\pm$ 0.29 & 1.64 $\pm$ 0.31 & 2.09 $\pm$ 0.42 & 1.98 $\pm$ 0.31 \\

\bottomrule
\end{tabular}
\end{table*}

\begin{figure*}[t]
    \centering

    \begin{subfigure}{\textwidth}
        \centering
        \includegraphics[width=\textwidth]{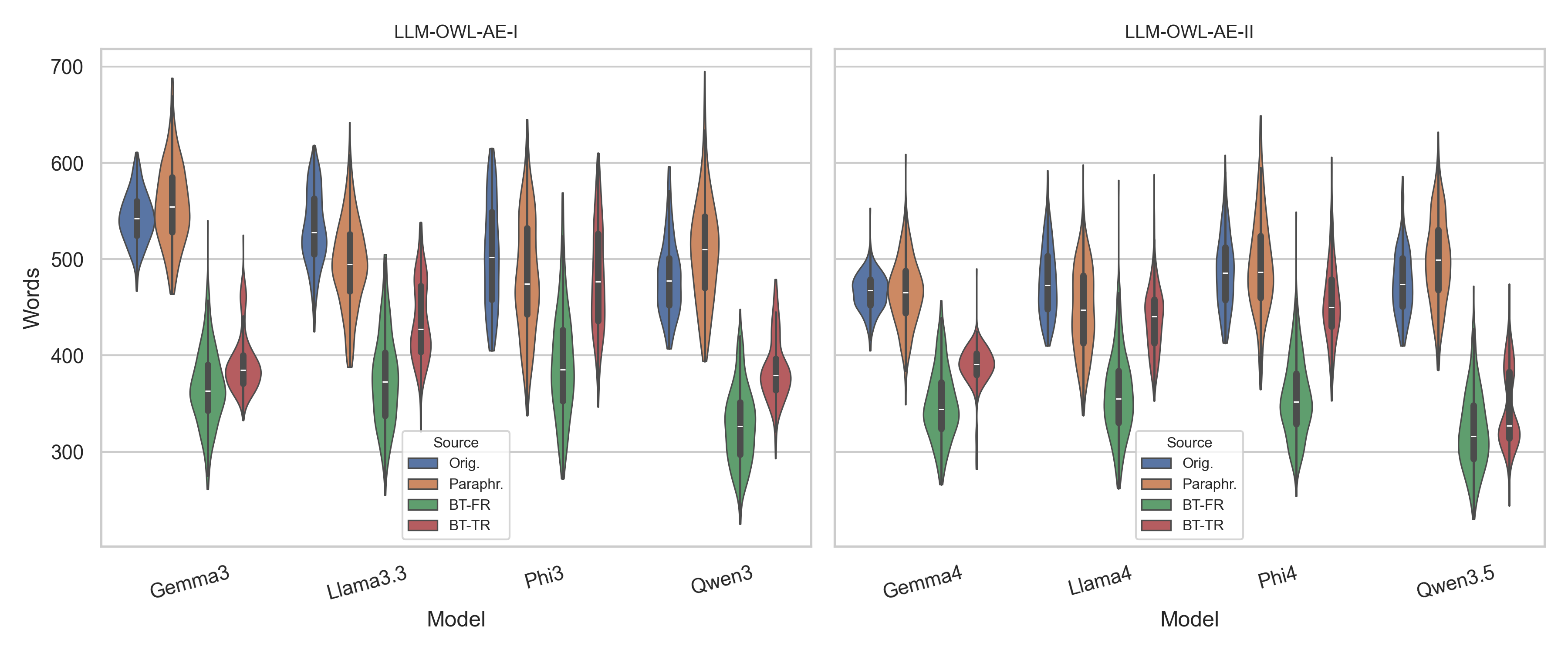}
        \caption{Mean number of words per essay.}
        \label{fig:feat_words}
    \end{subfigure}

    \vspace{0.5em}

    \begin{subfigure}{\textwidth}
        \centering
        \includegraphics[width=\textwidth]{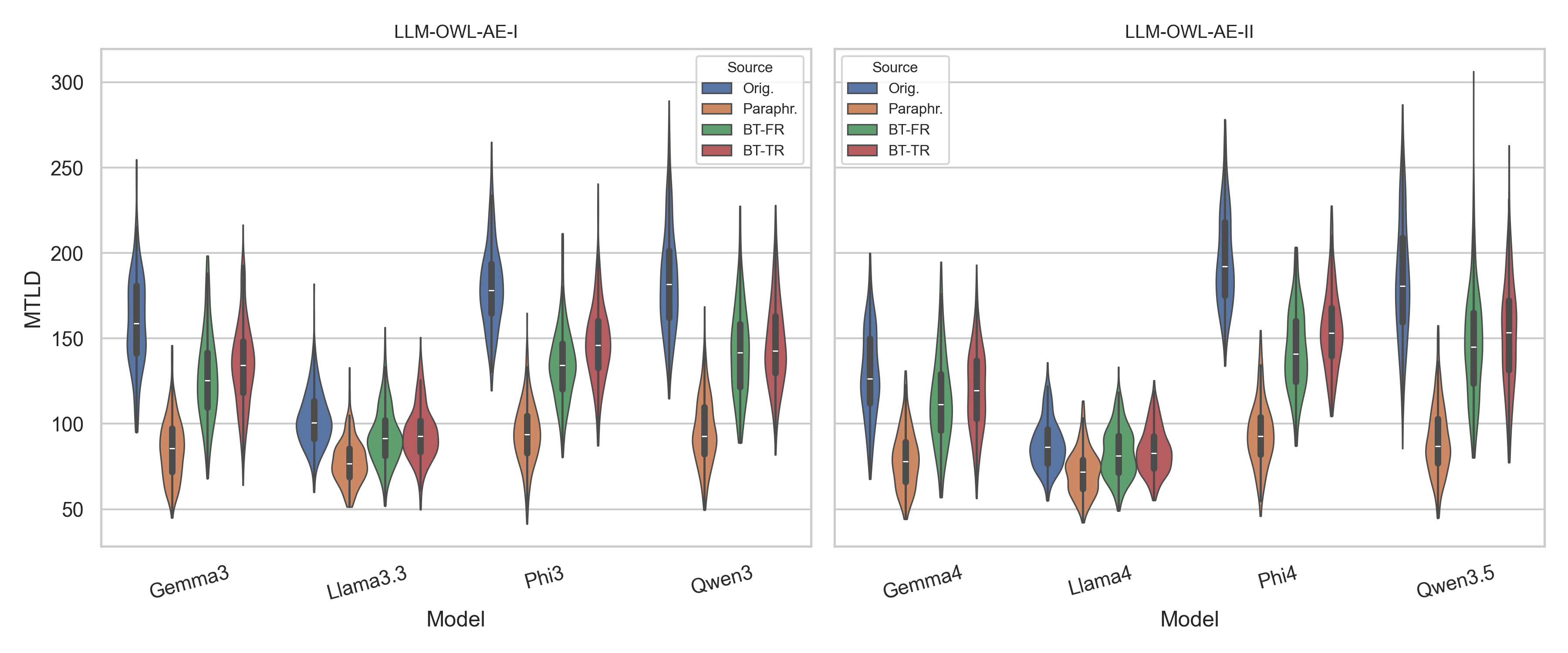}
        \caption{Mean MTLD score per essay.}
        \label{fig:feat_mtld}
    \end{subfigure}

    \vspace{0.5em}

    \begin{subfigure}{\textwidth}
        \centering
        \includegraphics[width=\textwidth]{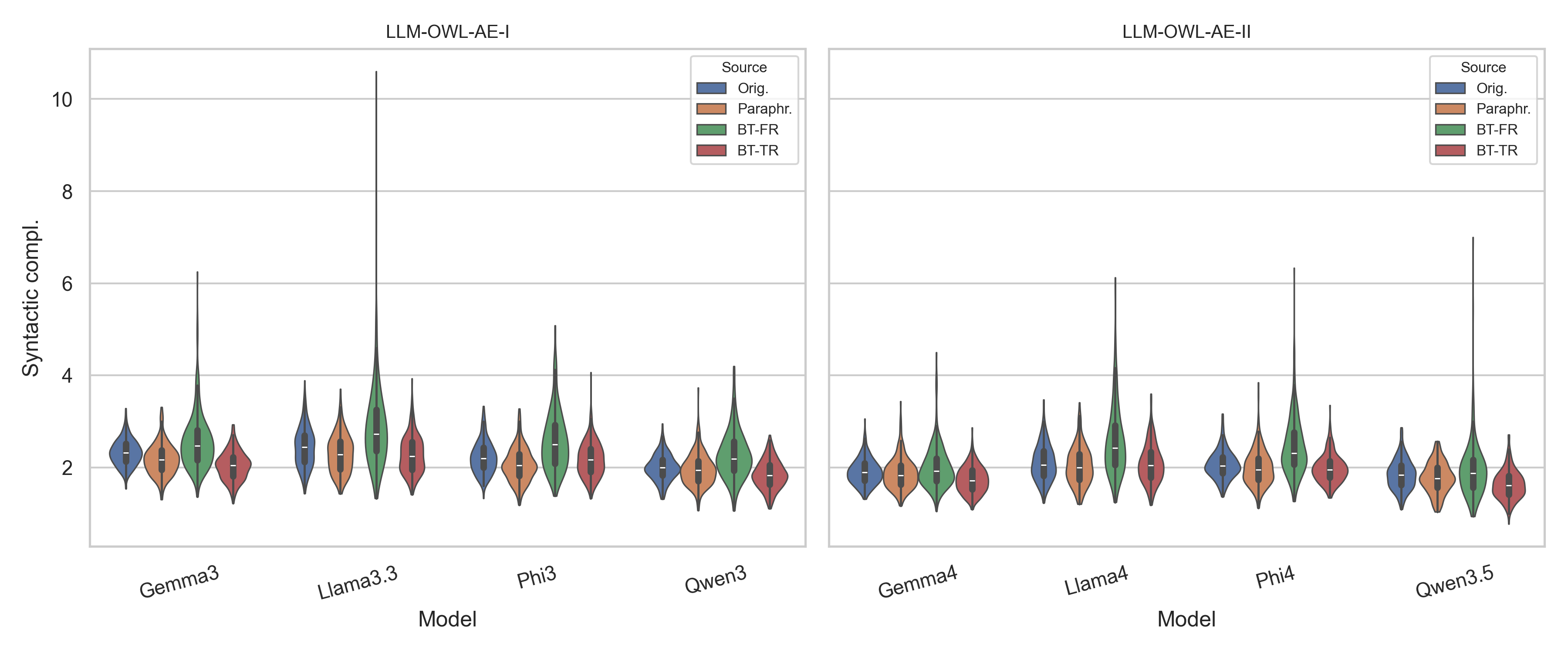}
        \caption{Mean syntactic complexity score per sentence per essay.}
        \label{fig:feat_syntcomp}
    \end{subfigure}

    \caption{Essay linguistic features. Violin plots on the left are for the LLM-OWL-AE-I data partition; violin plots on the right for LLM-OWL-AE-II data partition.}
    \label{fig:all_features}
\end{figure*}

\clearpage
\newpage

\section{Reasoning Graph Statistics}
\label{app:graph_stats}

\begin{table*}[h]
\small
\centering
\caption{Average number of nodes and edges per essay across different obfuscation strategies and thresholds.}
\label{tab:arg-features}
\begin{tabular}{lllcccccccc}
\toprule
& &
& \multicolumn{4}{c}{LLM-OWL-AE-I}
& \multicolumn{4}{c}{LLM-OWL-AE-II} \\
\cmidrule(lr){4-7} \cmidrule(lr){8-11}

Obfus. & Prop. & Thresh.
& Gemma3 & Qwen3 & Llama3 & Phi3
& Gemma4 & Qwen3.5 & Llama4 & Phi4 \\
\midrule

\multirow{5}{*}{Orig.} 
& Nodes & -- & 25.08 & 23.63 & 21.79 & 20.83 & 20.4 & 23.34 & 21.6 & 22.84 \\
& \multirow{4}{*}{Edges} & None & 304.35 & 270.24 & 228.98 & 210.58 & 199.05 & 264.68 & 225.58 & 252.33 \\
&  & $T_1$ & 258.24 & 242.52 & 214.98 & 191.92 & 176.82 & 222.01 & 210.7 & 229.61 \\
&  & $T_2$ & 225.21 & 215.02 & 188.03 & 169.42 & 157.85 & 199.29 & 180.89 & 204.88 \\
&  & $T_3$ & 171.68 & 169.13 & 149.37 & 133.52 & 125.25 & 160.58 & 142.74 & 163.39 \\
\midrule

\multirow{5}{*}{Paraphr.}
& Nodes & --                    & 23.75 & 22.1 & 20.75 & 19.59 & 19.49 & 22.1 & 20.36 & 21.42 \\
& \multirow{4}{*}{Edges}& None  & 273.37 & 236.9 & 208.19 & 185.9 & 182.26 & 237.18 & 200.88 & 222.79 \\
&  & $T_1$                      & 216.78 & 198.88 & 184.1 & 159.57 & 152.13 & 186.31 & 176.91 & 189.02 \\
&  & $T_2$                      & 187.85 & 173.89 & 158.82 & 138.28 & 133.52 & 164.78 & 149.84 & 165.07 \\
&  & $T_3$                      & 139.93 & 132.86 & 122.98 & 106.32 & 102.45 & 128.40 & 115.26 & 127.41 \\
\midrule

\multirow{5}{*}{BT-FR}
& Nodes & --                    & 15.24 & 14.18 & 13.62 & 13.72 & 14.68 & 14.65 & 14.09 & 14.23 \\
& \multirow{4}{*}{Edges}& None  & 111.35 & 96.52 & 89.80 & 90.88 & 102.80 & 104.42 & 95.40 & 97.25 \\
&  & $T_1$                      & 92.14 & 84.33 & 83.89 & 82.85 & 87.74 & 83.83 & 88.7 & 87.84 \\
&  & $T_2$                      & 79.86 & 73.79 & 72.21 & 72.37 & 77.05 & 74.16 & 74.11 & 77.45 \\
&  & $T_3$                      & 59.81 & 55.50 & 54.44 & 55.33 & 58.71 & 57.62 & 56.01 & 59.25 \\
\midrule

\multirow{5}{*}{BT-TR}
& Nodes & --                    & 19.16 & 19.31 & 19.17 & 19.96 & 18.24 & 17.83 & 20.31 & 21.56 \\
& \multirow{4}{*}{Edges}& None  & 176.72 & 179.60 & 177.28 & 193.29 & 158.64 & 155.13 & 199.43 & 224.96 \\
&  & $T_1$                      & 137.17 & 151.09 & 161.11 & 170.1 & 130.36 & 118.93 & 181.84 & 195.52 \\
&  & $T_2$                      & 115.01 & 129.34 & 137.5 & 148.35 & 112.3 & 103.14 & 154.22 & 170.63 \\
&  & $T_3$                      & 83.64 & 97.67 & 105.50 & 115.92 & 84.73 & 78.70 & 120.31 & 131.58 \\
\bottomrule
\end{tabular}
\end{table*}

\vspace{2em}

\begin{figure*}[h]
    \centering

    \begin{subfigure}{0.34\textwidth}
        \centering
        \includegraphics[width=\linewidth]{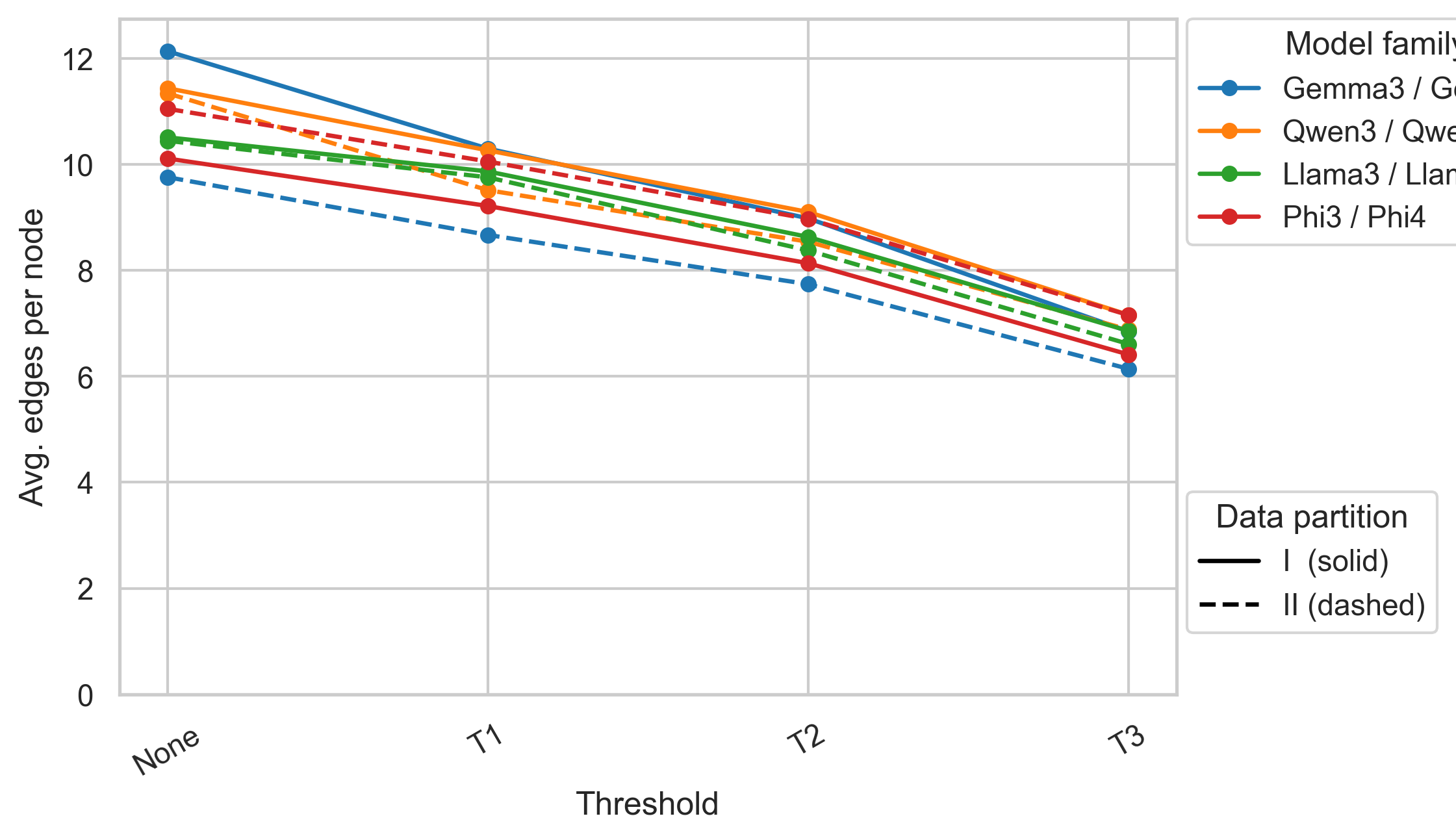}
        \caption{Original essays.}
        \label{fig:edges_orig}
    \end{subfigure}
    \hspace{0.02\textwidth}
    \begin{subfigure}{0.34\textwidth}
        \centering
        \includegraphics[width=\linewidth]{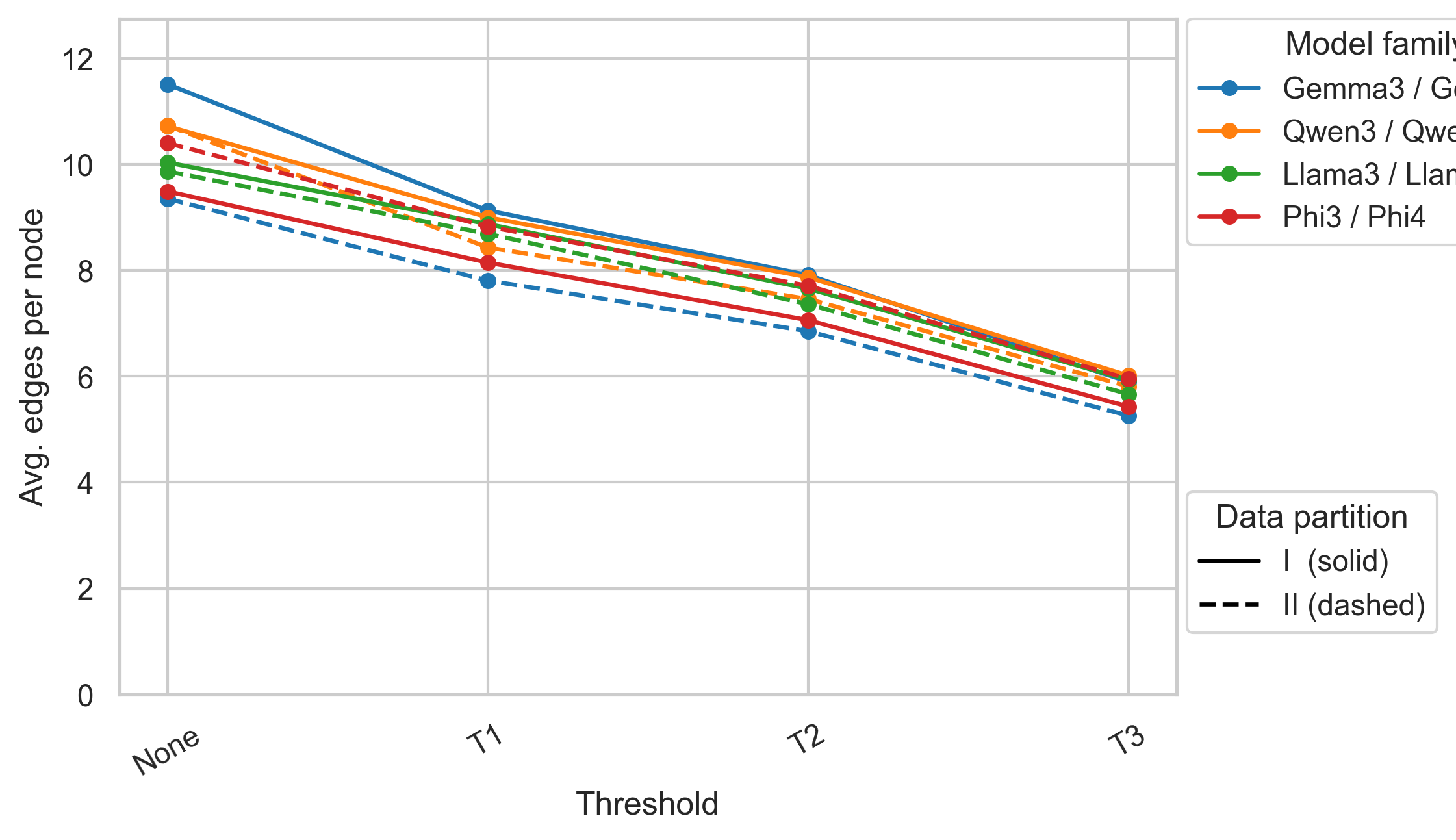}
        \caption{Paraphrased essays.}
        \label{fig:edges_para}
    \end{subfigure}

    \vspace{0.1em}

    \begin{subfigure}{0.34\textwidth}
        \centering
        \includegraphics[width=\linewidth]{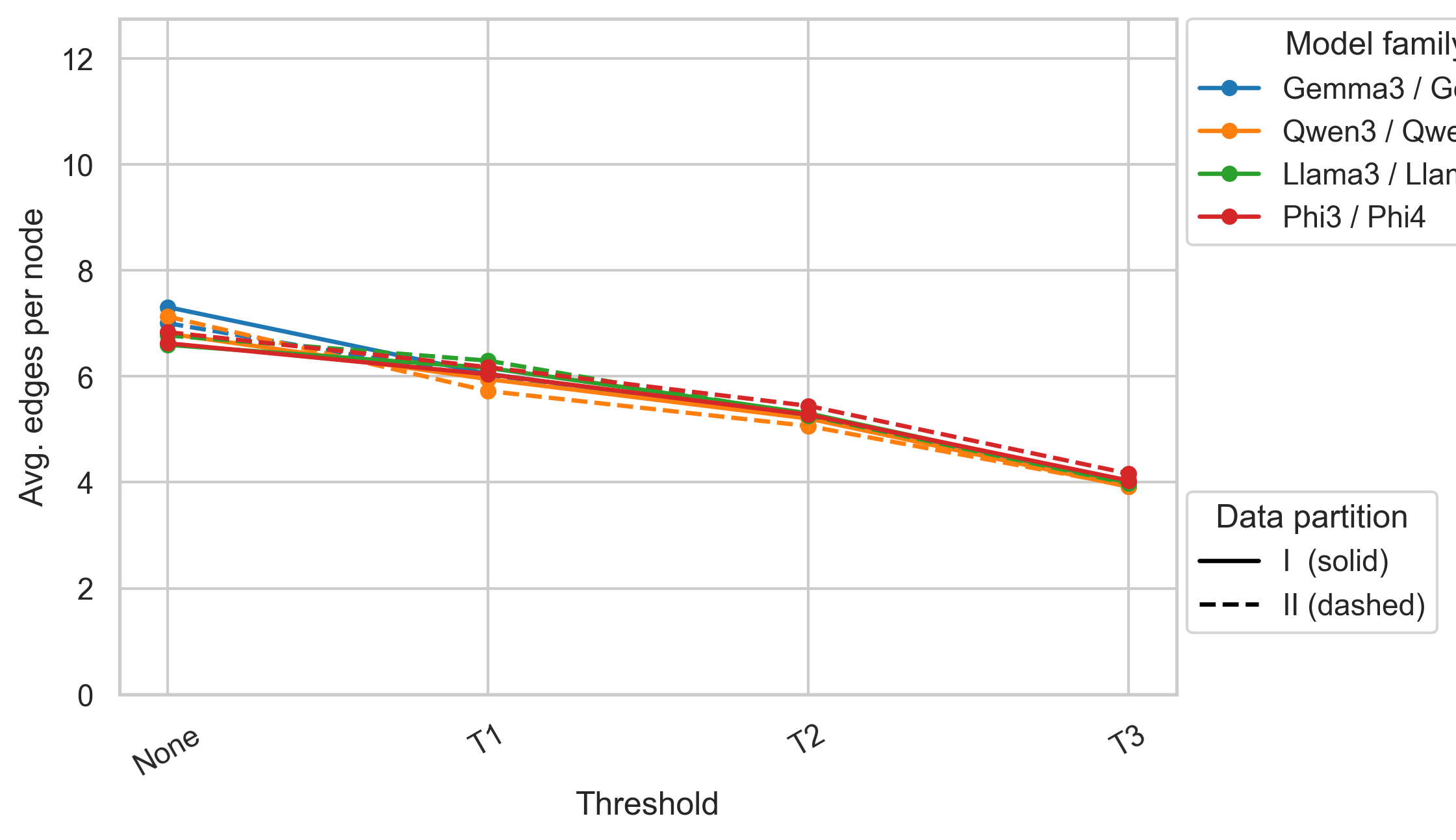}
        \caption{Backtranslated essays (French).}
        \label{fig:edges_btfr}
    \end{subfigure}
    \hspace{0.02\textwidth}
    \begin{subfigure}{0.34\textwidth}
        \centering
        \includegraphics[width=\linewidth]{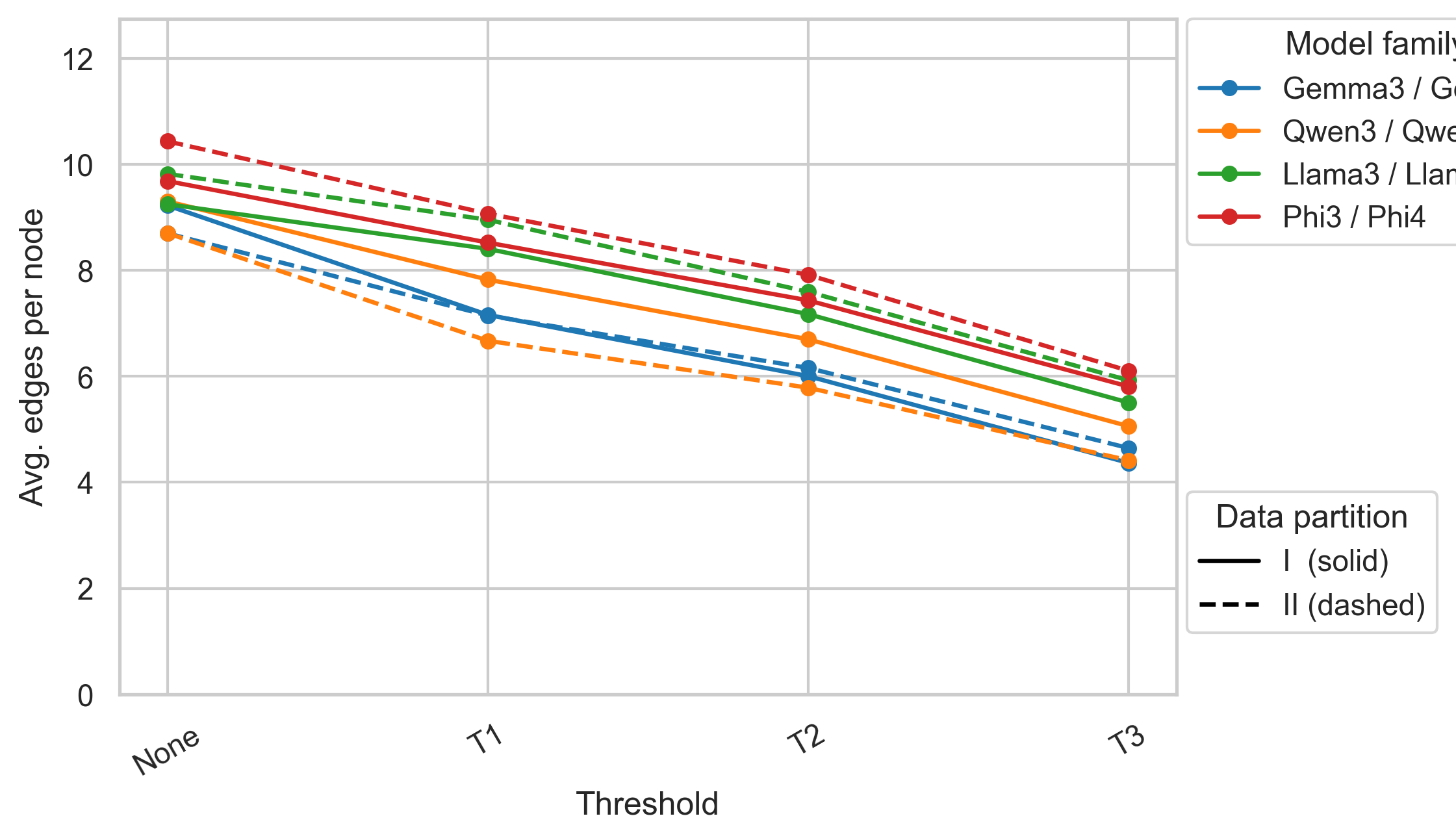}
        \caption{Backtranslated essays (Turkish)}
        \label{fig:edges_bttr}
    \end{subfigure}

    \caption{Ratio of the average number of edges per node across different thresholds.}
    \label{fig:all_figures}
\end{figure*}

\begin{figure*}[h]
    \centering

    \begin{subfigure}{0.48\textwidth}
        \centering
        \includegraphics[width=\linewidth]{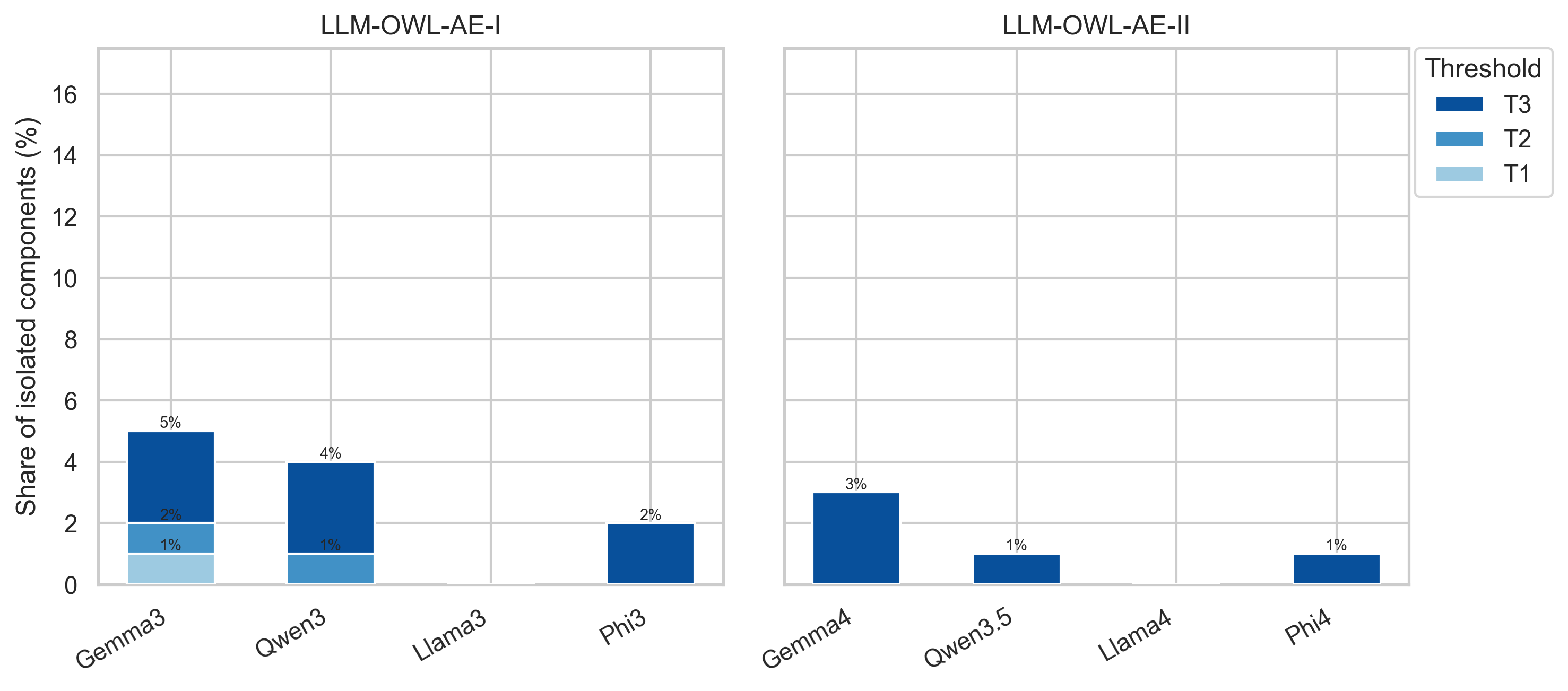}
        \caption{Original essays.}
        \label{fig:isol_orig}
    \end{subfigure}
    \hfill
    \begin{subfigure}{0.48\textwidth}
        \centering
        \includegraphics[width=\linewidth]{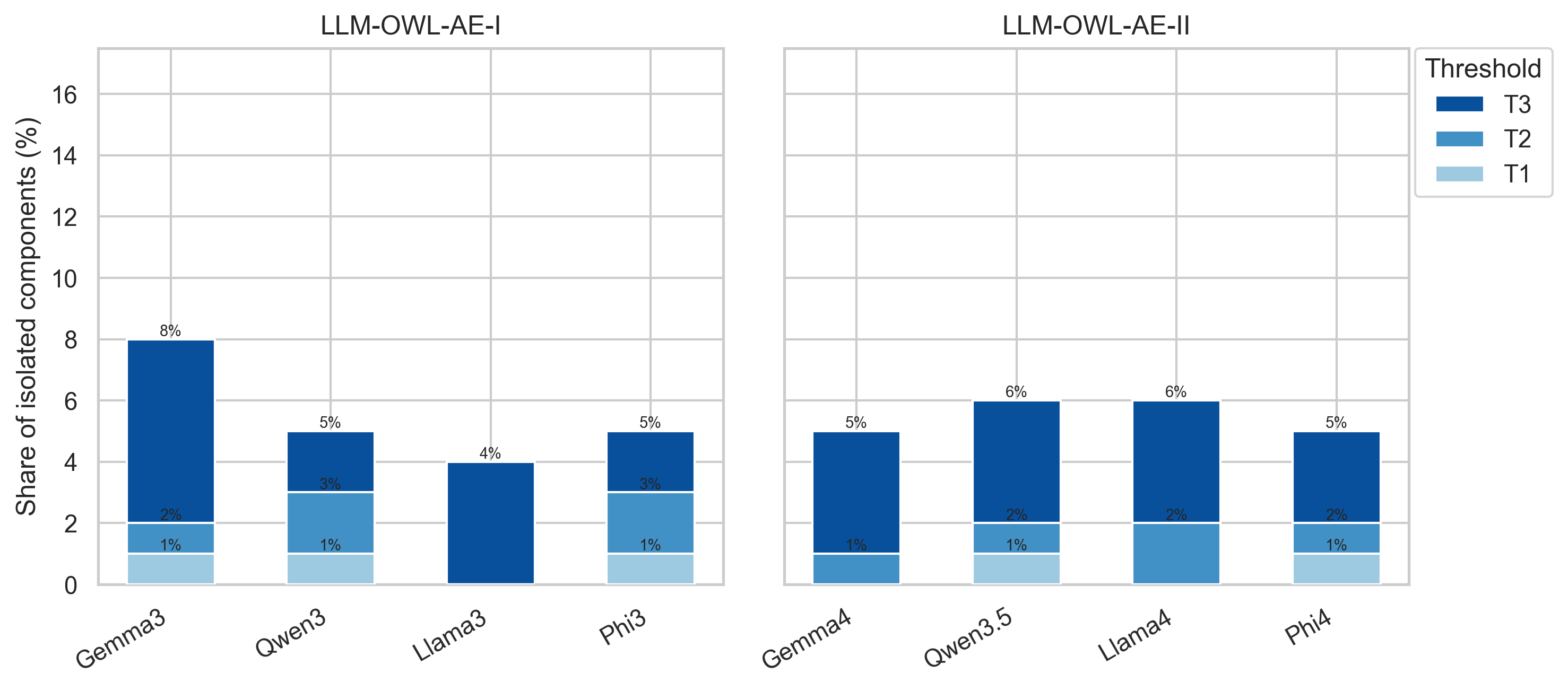}
        \caption{Paraphrased essays.}
        \label{fig:isol_para}
    \end{subfigure}

    \vspace{0.5em}

    \begin{subfigure}{0.48\textwidth}
        \centering
        \includegraphics[width=\linewidth]{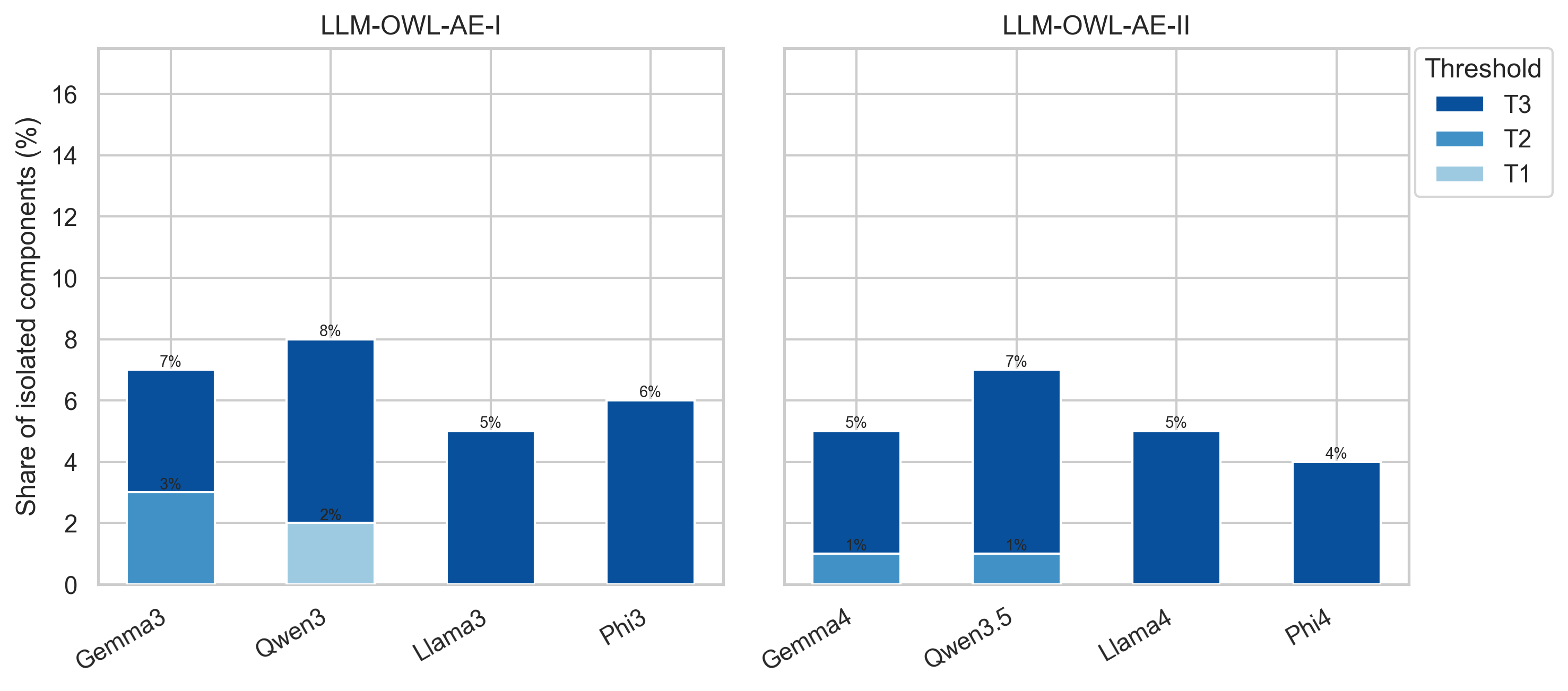}
        \caption{Backtranslated essays (French).}
        \label{fig:isol_btfr}
    \end{subfigure}
    \hfill
    \begin{subfigure}{0.48\textwidth}
        \centering
        \includegraphics[width=\linewidth]{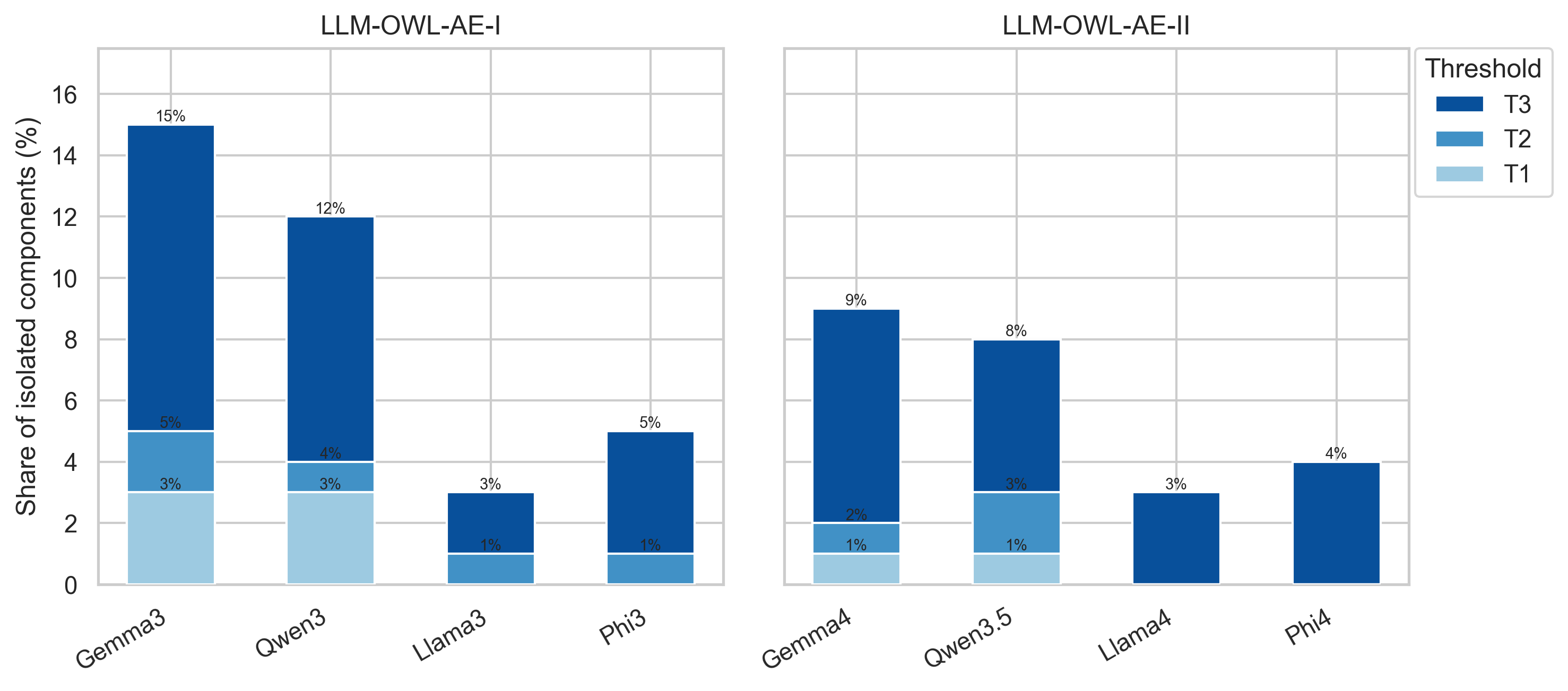}
        \caption{Backtranslated essays (Turkish)}
        \label{fig:isol_bttr}
    \end{subfigure}

    \caption{Proportion of isolated components in graphs.}
    \label{fig:all_figures}
\end{figure*}

\begin{figure*}[h]
    \centering

    \begin{subfigure}{0.48\textwidth}
        \centering
        \includegraphics[width=\linewidth]{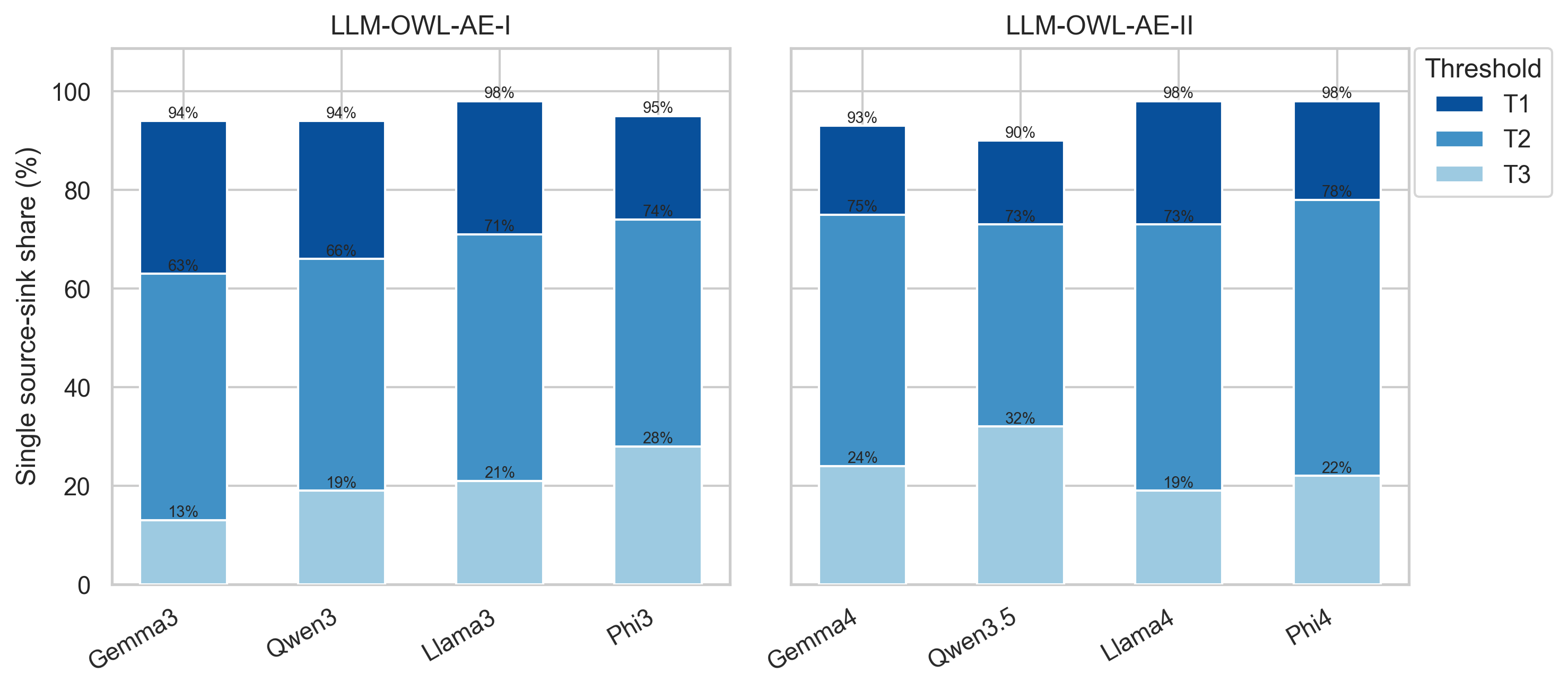}
        \caption{Original essays.}
        \label{fig:sink_orig}
    \end{subfigure}
    \hfill
    \begin{subfigure}{0.48\textwidth}
        \centering
        \includegraphics[width=\linewidth]{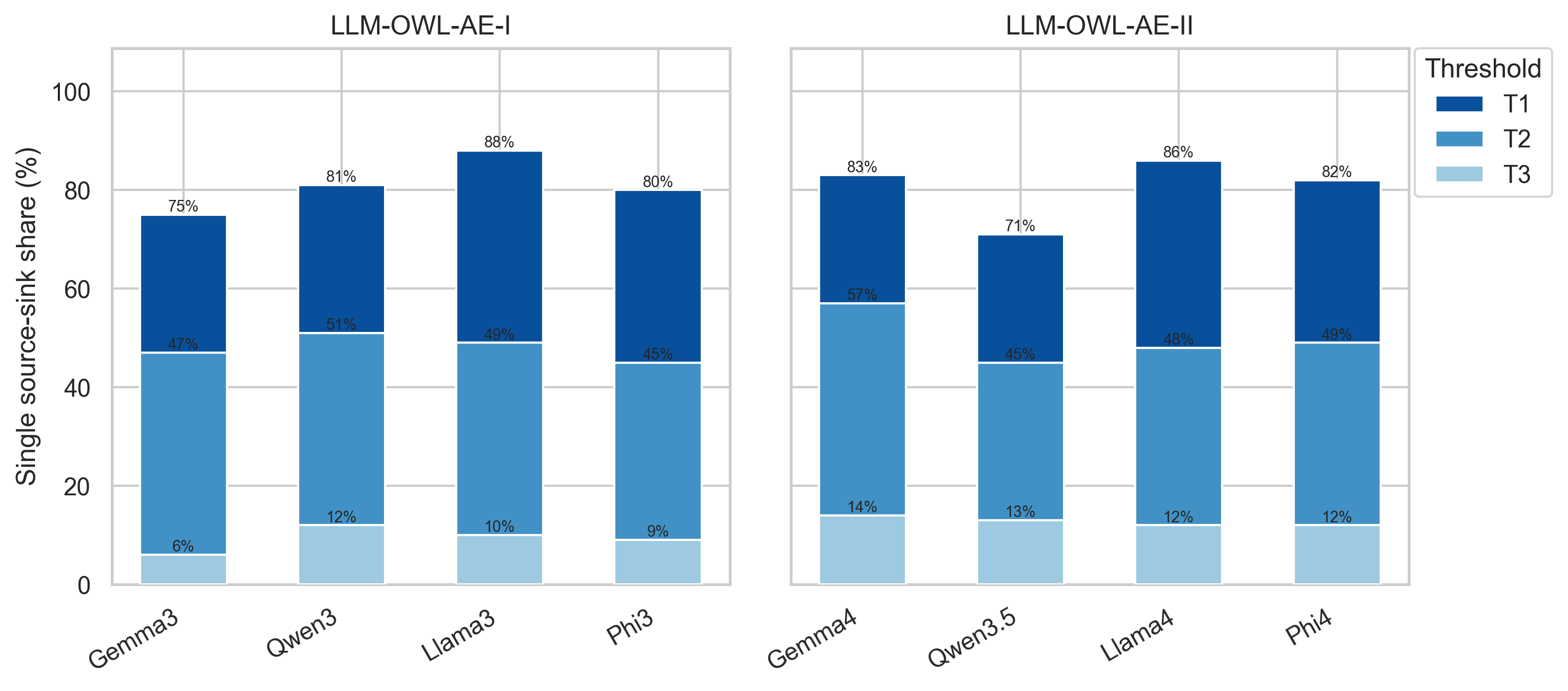}
        \caption{Paraphrased essays.}
        \label{fig:sink_para}
    \end{subfigure}

    \vspace{0.5em}

    \begin{subfigure}{0.48\textwidth}
        \centering
        \includegraphics[width=\linewidth]{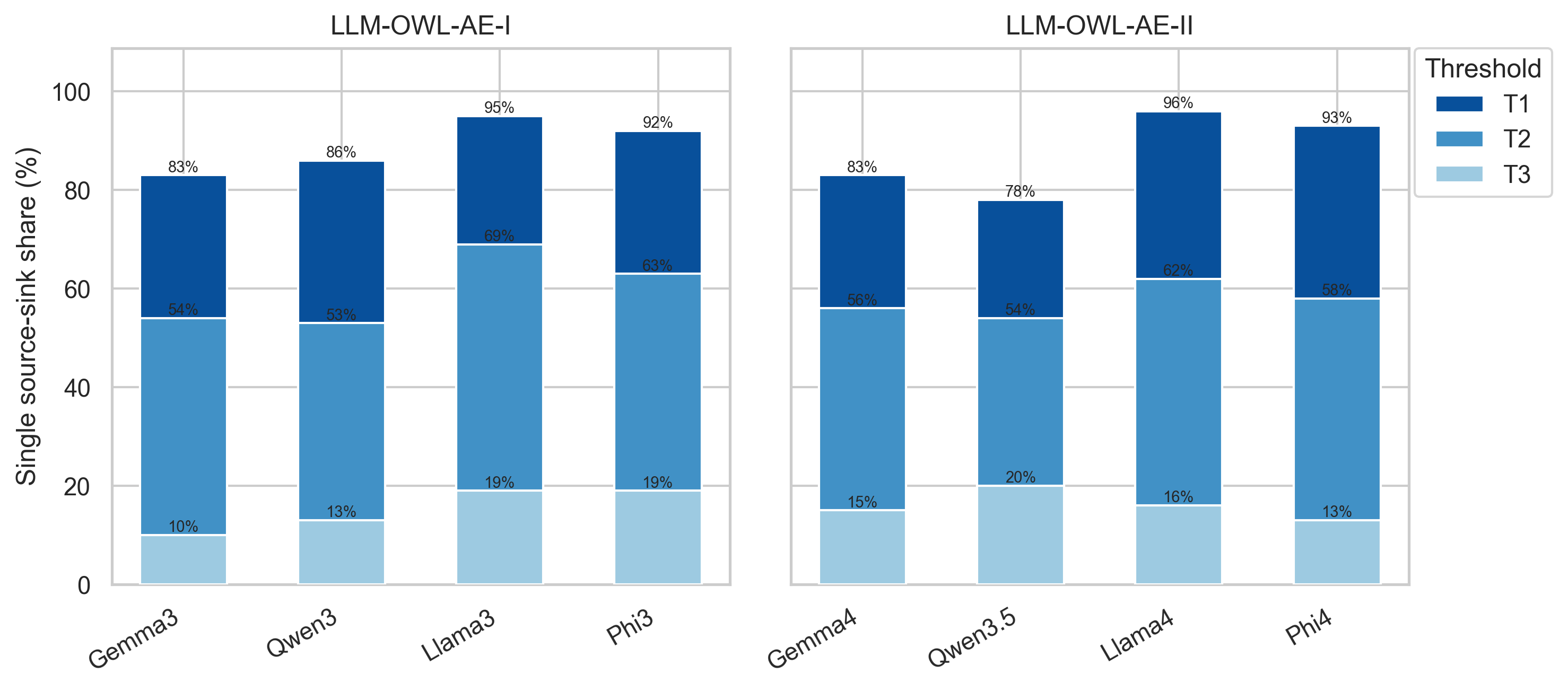}
        \caption{Backtranslated essays (French).}
        \label{fig:sink_btfr}
    \end{subfigure}
    \hfill
    \begin{subfigure}{0.48\textwidth}
        \centering
        \includegraphics[width=\linewidth]{figures/single_source_sink_bt-fr.png}
        \caption{Backtranslated essays (Turkish)}
        \label{fig:sink_bttr}
    \end{subfigure}

    \caption{Proportion of the graphs forming a single source-sink path.}
    \label{fig:all_figures}
\end{figure*}

\clearpage
\section{Extended Experimental Results}
Extended experimental results with all threshold configurations for GNN architectures are included. While we evaluate 1- to 7-layer GNN architectures, the best results were consistently achieved with 1-3 layers, so we do not report results for models with more than three layers. 
\begin{table*}[h]
\small
\centering
\caption{Macro F1-score results in the \textsc{Same-Version} setup, $\ast$ is used to indicate when the best result is achieved with a 2-layer GNN, $\dagger$ is used for a 3-layer GNN, and the rest of the results are reported for a 1-layer GNN. \textit{None} in Threshold column indicates a complete graph. \textit{Argmax} and \textit{probs} refer to the edge processing strategies defined in Section \ref{sec:method}. `Orig.' refers to the non-obfuscated essays, `paraphr.' to the paraphrased, `BT-FR' and `BT-TR' to the backtranslated essays via French and Turkish, respectively.}
\label{tab:samemodel-versions-layers1}

\begin{tabular}{lllcccccccc}
\toprule
& &
& \multicolumn{4}{c}{LLM-OWL-AE-I}
& \multicolumn{4}{c}{LLM-OWL-AE-II} \\
\cmidrule(lr){4-7} \cmidrule(lr){8-11}

Model & Edge & Thresh.
& Orig. & Paraphr. & BT-FR & BT-TR
& Orig. & Paraphr. & BT-FR & BT-TR \\
\midrule

\multicolumn{11}{l}{\textit{Text-only baseline}} \\
Longformer          & -- & -- & 0.97 & 0.63 & 0.84 & 0.88 & 0.96 & 0.45 & 0.62 & 0.69 \\
\midrule

\multicolumn{11}{l}{\textit{Text + Reasoning Structure}} \\
\multirow[t]{8}{*}{GCN} & \multirow[t]{4}{*}{argmax} & None & 0.66 & 0.56 & 0.57 & 0.61 & 0.68 & 0.59 & 0.61 & 0.64 \\
 &  & $T_1$ & 0.67 & 0.52 & 0.55 & 0.62 & 0.69 & 0.61 & 0.62 & 0.64 \\
 &  & $T_2$ & 0.69 & 0.51 & 0.58 & 0.62 & 0.70 & 0.60 & 0.61 & 0.65 \\
 &  & $T_3$ & 0.67 & 0.59 & 0.58 & 0.66 & 0.72 & 0.63 & 0.59 & 0.64 \\
 & \multirow[t]{4}{*}{probs} & None & 0.66 & 0.56 & 0.57 & 0.61 & 0.68 & 0.59 & 0.61 & 0.64 \\
 &  & $T_1$ & 0.67 & 0.52 & 0.55 & 0.62 & 0.69 & 0.61 & 0.62 & 0.65$\ast$ \\
 &  & $T_2$ & 0.67 & 0.50 & 0.55 & 0.59 & 0.70 & 0.60 & 0.61 & 0.65 \\
 &  & $T_3$ & 0.67 & 0.59 & 0.58 & 0.66 & 0.72 & 0.63 & 0.59 & 0.64 \\
\midrule
\multirow[t]{8}{*}{GAT} & \multirow[t]{4}{*}{argmax} & None & 0.73 & 0.60 & 0.67 & 0.66 & 0.75 & 0.61 & 0.68 & 0.70 \\
 &  & $T_1$ & 0.71 & 0.59 & 0.64 & 0.66 & 0.72 & 0.63 & 0.66 & 0.68 \\
 &  & $T_2$ & 0.70 & 0.58 & 0.65 & 0.66 & 0.75 & 0.64 & 0.67 & 0.70 \\
 &  & $T_3$ & 0.70 & 0.59 & 0.64 & 0.68 & 0.73 & 0.61 & 0.65 & 0.68 \\
 & \multirow[t]{4}{*}{probs} & None & 0.74 & 0.62 & 0.66 & 0.67 & 0.77 & 0.63 & 0.67 & 0.72 \\
 &  & $T_1$ & 0.69 & 0.58 & 0.64 & 0.65 & 0.72 & 0.60 & 0.65 & 0.69 \\
 &  & $T_2$ & 0.69$\ast$ & 0.58 & 0.63 & 0.64 & 0.76 & 0.60 & 0.65 & 0.70 \\
 &  & $T_3$ & 0.67 & 0.63 & 0.65 & 0.66 & 0.73 & 0.61 & 0.63 & 0.69 \\
\midrule
\multirow[t]{8}{*}{Graph Transformer} & \multirow[t]{4}{*}{argmax} & None & 0.73 & 0.62 & 0.63 & 0.67 & 0.78 & 0.69 & 0.69 & 0.72 \\
 &  & $T_1$ & 0.70 & 0.58 & 0.65$\ast$ & 0.66$\ast$ & 0.77 & 0.71 & 0.68 & 0.70 \\
 &  & $T_2$ & 0.72$\ast$ & 0.59 & 0.62$\ast$ & 0.64$\ast$ & 0.77 & 0.69 & 0.67 & 0.71 \\
 &  & $T_3$ & 0.69 & 0.55 & 0.57$\ast$ & 0.61$\ast$ & 0.77 & 0.69 & 0.66 & 0.70 \\
 & \multirow[t]{4}{*}{probs} & None & 0.74 & 0.63 & 0.67 & 0.69 & 0.81 & 0.72 & 0.72 & 0.70 \\
 &  & $T_1$ & 0.71 & 0.60$\ast$ & 0.64$\ast$ & 0.65 & 0.80 & 0.68$\ast$ & 0.68 & 0.71 \\
 &  & $T_2$ & 0.70 & 0.58 & 0.60 & 0.64 & 0.78 & 0.63 & 0.68 & 0.71 \\
 &  & $T_3$ & 0.69 & 0.56 & 0.59 & 0.62 & 0.78 & 0.68 & 0.65 & 0.69 \\
\midrule
\multirow[t]{8}{*}{GPS} & \multirow[t]{4}{*}{argmax} & None & 0.74$\ast$ & 0.60 & 0.60 & 0.63 & 0.78$\ast$ & 0.71$\ast$ & 0.60 & 0.65 \\
 &  & $T_1$ & 0.75$\ast$ & 0.59$\ast$ & 0.59 & 0.61 & 0.76 & 0.68$\ast$ & 0.59 & 0.66 \\
 &  & $T_2$ & 0.77 & 0.59 & 0.61 & 0.67 & 0.76$\ast$ & 0.67$\dagger$ & 0.66 & 0.67 \\
 &  & $T_3$ & 0.75 & 0.62 & 0.67 & 0.67 & 0.76 & 0.66 & 0.65 & 0.71 \\
 & \multirow[t]{4}{*}{probs} & None & 0.74$\dagger$ & 0.64$\ast$ & 0.61 & 0.66 & 0.80$\ast$ & 0.69$\dagger$ & 0.60 & 0.66 \\
 &  & $T_1$ & 0.77 & 0.59 & 0.62 & 0.67 & 0.76 & 0.67$\ast$ & 0.68 & 0.70 \\
 &  & $T_2$ & 0.76 & 0.60 & 0.61 & 0.67 & 0.78$\dagger$ & 0.66 & 0.65 & 0.68 \\
 &  & $T_3$ & 0.73 & 0.60$\ast$ & 0.63 & 0.65 & 0.77$\ast$ & 0.69$\dagger$ & 0.65 & 0.67 \\
\bottomrule

\end{tabular}
\end{table*}

\begin{table*}[t]
\small
\centering
\caption{Macro F1-score results in the \textsc{Cross-Version} setup for 1-layer GNN configurations, $\ast$ is used to indicate when the best result is achieved with a 2-layer GNN, $\dagger$ is used for a 3-layer GNN, and the rest of the results are reported for a 1-layer GNN. \textit{None} in Threshold column indicates a complete graph. \textit{Argmax} and \textit{probs} refer to the edge processing strategies defined in Section \ref{sec:method}. `Orig.' refers to the non-obfuscated essays, `paraphr.' to the paraphrased, `BT-FR' and `BT-TR' to the backtranslated essays via French and Turkish, respectively.}
\label{tab:diffmodel-versions-layers1}

\begin{tabular}{lcccccccccc}
\toprule
& & &
\multicolumn{4}{c}{\makecell{LLM-OWL-AE-I train /\\ LLM-OWL-AE-II test}}
& \multicolumn{4}{c}{\makecell{LLM-OWL-AE-II train /\\ LLM-OWL-AE-I test}} \\
\cmidrule(lr){4-7} \cmidrule(lr){8-11}

Model & Edge & Thresh. 
& Orig. & Paraphr. & BT-FR & BT-TR
& Orig. & Paraphr. & BT-FR & BT-TR \\
\midrule

\multicolumn{11}{l}{\textit{Text-only baseline}} \\
Longformer         & -- & -- & 0.47 & 0.30 & 0.56 & 0.50 & 0.49 & 0.36 & 0.35 & 0.37 \\
\midrule

\multicolumn{11}{l}{\textit{Text + Reasoning Structure}} \\
\multirow[t]{8}{*}{GCN} & \multirow[t]{4}{*}{argmax} & None & 0.57 & 0.46 & 0.56 & 0.55 & 0.50 & 0.42 & 0.43 & 0.49 \\
 &  & $T_1$ & 0.60 & 0.46 & 0.56 & 0.56 & 0.49 & 0.41 & 0.42 & 0.46 \\
 &  & $T_2$ & 0.59 & 0.46 & 0.56 & 0.57 & 0.46 & 0.40$\ast$ & 0.40$\ast$ & 0.45$\ast$ \\
 &  & $T_3$ & 0.58 & 0.48 & 0.55 & 0.55 & 0.48 & 0.40 & 0.39 & 0.46 \\
 & \multirow[t]{4}{*}{probs} & None & 0.57 & 0.46 & 0.55 & 0.55 & 0.50 & 0.42 & 0.43 & 0.49 \\
 &  & $T_1$ & 0.60 & 0.46 & 0.56 & 0.56 & 0.49 & 0.41 & 0.42 & 0.46 \\
 &  & $T_2$ & 0.60 & 0.46 & 0.57 & 0.58 & 0.46 & 0.40$\ast$ & 0.40$\ast$ & 0.45$\ast$ \\
 &  & $T_3$ & 0.58 & 0.48 & 0.55 & 0.55 & 0.48 & 0.40 & 0.39 & 0.47 \\
\midrule
\multirow[t]{8}{*}{GAT} & \multirow[t]{4}{*}{argmax} & None & 0.62 & 0.50 & 0.61 & 0.61 & 0.51 & 0.43 & 0.42 & 0.50 \\
 &  & $T_1$ & 0.64 & 0.53 & 0.61 & 0.61 & 0.53 & 0.44 & 0.45 & 0.50 \\
 &  & $T_2$ & 0.60 & 0.51 & 0.58 & 0.58 & 0.50 & 0.41 & 0.42 & 0.48 \\
 &  & $T_3$ & 0.61 & 0.53 & 0.58 & 0.59 & 0.47 & 0.38$\ast$ & 0.39$\dagger$ & 0.43 \\
 & \multirow[t]{4}{*}{probs} & None & 0.63 & 0.50 & 0.60 & 0.59 & 0.51 & 0.43$\ast$ & 0.43$\ast$ & 0.49 \\
 &  & $T_1$ & 0.62 & 0.49 & 0.59 & 0.58 & 0.47 & 0.39 & 0.40$\dagger$ & 0.46 \\
 &  & $T_2$ & 0.63 & 0.52 & 0.60 & 0.59 & 0.50 & 0.40 & 0.42 & 0.48 \\
 &  & $T_3$ & 0.59 & 0.51 & 0.60 & 0.57 & 0.53 & 0.42 & 0.44 & 0.50 \\
\midrule
\multirow[t]{8}{*}{Graph Transformer} & \multirow[t]{4}{*}{argmax} & None & 0.65 & 0.54 & 0.64 & 0.63 & 0.52 & 0.45 & 0.44 & 0.49 \\
 &  & $T_1$ & 0.63 & 0.52 & 0.60 & 0.60 & 0.51 & 0.45 & 0.44 & 0.48 \\
 &  & $T_2$ & 0.63 & 0.52 & 0.60 & 0.61 & 0.49 & 0.42 & 0.42 & 0.48 \\
 &  & $T_3$ & 0.66 & 0.51 & 0.59 & 0.62 & 0.51 & 0.42 & 0.43 & 0.49 \\
 & \multirow[t]{4}{*}{probs} & None & 0.59 & 0.46 & 0.59 & 0.54 & 0.51 & 0.47 & 0.46 & 0.53 \\
 &  & $T_1$ & 0.64 & 0.53 & 0.61 & 0.61 & 0.53 & 0.42 & 0.44 & 0.49 \\
 &  & $T_2$ & 0.66 & 0.52 & 0.61 & 0.63 & 0.50 & 0.41$\ast$ & 0.42 & 0.48 \\
 &  & $T_3$ & 0.63 & 0.49 & 0.60 & 0.60 & 0.50 & 0.42 & 0.43 & 0.48 \\
\midrule
\multirow[t]{8}{*}{GPS} & \multirow[t]{4}{*}{argmax} & None & 0.59 & 0.49 & 0.56 & 0.57 & 0.53 & 0.43 & 0.46 & 0.53 \\
 &  & $T_1$ & 0.64 & 0.51 & 0.61 & 0.61 & 0.50 & 0.41 & 0.43 & 0.55$\ast$ \\
 &  & $T_2$ & 0.64 & 0.53 & 0.60 & 0.61 & 0.48 & 0.41 & 0.48 & 0.55$\ast$ \\
 &  & $T_3$ & 0.63 & 0.52 & 0.62 & 0.60 & 0.50$\ast$ & 0.43 & 0.47 & 0.55$\ast$ \\
 & \multirow[t]{4}{*}{probs} & None & 0.59 & 0.49 & 0.55 & 0.58 & 0.53 & 0.46 & 0.46 & 0.57 \\
 &  & $T_1$ & 0.64 & 0.51 & 0.60 & 0.62 & 0.53 & 0.43 & 0.47 & 0.56$\ast$ \\
 &  & $T_2$ & 0.63 & 0.51 & 0.59 & 0.59 & 0.51 & 0.41$\ast$ & 0.45 & 0.55 \\
 &  & $T_3$ & 0.63 & 0.52 & 0.60 & 0.60 & 0.52 & 0.43 & 0.45 & 0.53$\ast$ \\
\bottomrule

\end{tabular}
\end{table*}
\twocolumn
\section{Computational Resources}
For the data generation, we used the following GPU resources: NVIDIA GeForce RTX 4090 24 GB and NVIDIA GeForce RTX 5090 32 GB. The RTX 4090 was additionally used for the Longformer experiments. RTX 5090 was used for the oAMF argument mining pipeline (module ARIR) and GNN experiments. The other modules of the pipeline, the linguistic features, and the graph statistics were computed without GPU acceleration.

\section{Examples of Original and Obfuscated Essays}
We include an example of a paragraph from an essay generated by Llama3.3 on the topic of Animal Rights (stance: against) as well as its obfuscated versions. The texts are copied verbatim from the dataset. The examples demonstrate that the obfuscated paragraphs closely follow the original content in meaning and retain fluency, while exhibiting surface-level lexical and syntactic differences.

\paragraph{Original essay} \textit{In addition to the ethical concerns, animal testing is also unreliable and often produces false or misleading results. Animals have different physiological and biological systems than humans, which means that the results of animal tests may not be applicable to humans. For instance, a drug may be safe in animals but toxic in humans, or vice versa. This has led to many cases where drugs have been approved based on animal testing, only to be later withdrawn from the market due to adverse effects in humans. Moreover, alternative methods such as computer simulations, cell cultures, and human clinical trials are now available, which can provide more accurate and reliable results than animal testing.}
\paragraph{Paraphrased} \textit{Animals have different physiological and biological systems from humans, which means that the results of animal experiments may not be applicable to humans. A drug can be safe for animals but toxic for humans, or vice versa. In addition to being unreliable, animal experiments often produce misleading results. Besides, the methods of the computer simulation, the cell culture, and the clinical trial have all been developed. They are more reliable and accurate than animal testing. It is not uncommon for medicines approved based on animal experiments to later be withdrawn from the market because of side effects in humans.}
\paragraph{Backtranslated via French} \textit{In addition to ethical concerns, animal tests are also unreliable and often produce false or misleading results.The animals have physiological and biological systems different from those of humans, which means that the results of animal tests may not apply to humans. For example, a drug may be safe in animals but toxic in humans, or vice versa. This has led to many cases where drugs have been approved on the basis of animal tests, only to be removed later from the market due to adverse effects in humans.}
\paragraph{Backtranslated via Turkish} \textit{In addition to ethical concerns, animal testing is unreliable and often produces inaccurate or misleading results. Animals have different physiological and biological systems than humans, which means that the results of animal testing may not apply to humans. For example, a drug may be safe in animals, but it may be toxic in humans or vice versa. This has led to many cases where drugs are more accurate simulations than animal testing and subsequently withdrawn from the market due to adverse effects on human cells.}

\end{document}